\documentclass[11pt]{article}

\usepackage[preprint]{acl}

\usepackage{times}
\usepackage{geometry}
\usepackage{booktabs}
\usepackage{multirow}
\usepackage{array}
\usepackage{xcolor}
\usepackage{caption}
\usepackage{makecell}
\usepackage{graphicx} 
\usepackage[table]{xcolor}
\usepackage{latexsym}
\usepackage{amsmath, amssymb}
\usepackage{hyperref}
\usepackage{comment}
\usepackage{hhline}
\usepackage{tcolorbox}
\usepackage{enumitem}
\usepackage{booktabs}
\usepackage{float}
\usepackage{verbatimbox}
\usepackage{listings}
\usepackage{fvextra}

\usepackage[T1]{fontenc}

\usepackage{microtype}

\usepackage{inconsolata}

\usepackage{graphicx}

\usepackage{bbding}
\usepackage{enumitem}

\definecolor{taskbg}{HTML}{FFFCF6}      
\definecolor{taskborder}{RGB}{102, 102, 112}
\definecolor{inputbg}{RGB}{252, 247, 239}
\definecolor{outputbg}{RGB}{245, 249, 246}

\definecolor{tasktitlebg}{HTML}{EFEDE6}

\newtcolorbox{blackbox}[1]{
  colback=gray!10,       
  colframe=black!75,     
  coltitle=white,        
  fonttitle=\bfseries,   
  title=#1,              
  arc=3mm,               
  boxrule=0.5mm,         
  titlerule=0mm,         
  toptitle=2mm,          
  bottomtitle=2mm        
}

\tcbset{
    taskbox/.style={
        colback=taskbg,
        colframe=taskborder,
        colbacktitle=tasktitlebg,        
        coltitle=black,                  
        boxrule=1pt,
        arc=3pt,
        left=5pt,
        right=5pt,
        top=5pt,
        bottom=5pt,
        fonttitle=\bfseries,
    },
    examplebox/.style 2 args={
        colback=#1,
        colframe=taskborder,
        colbacktitle=tasktitlebg,       
        coltitle=black,                  
        boxrule=0.5pt,
        arc=2pt,
        left=4pt,
        right=4pt,
        top=4pt,
        bottom=4pt,
        fonttitle=\bfseries\small,
        title=#2,
    }
}

\hypersetup{
    colorlinks=true,
    linkcolor=red,
    citecolor=blue,
    filecolor=magenta,      
    urlcolor=blue,
linktocpage}

\usepackage{tcolorbox}
\tcbuselibrary{breakable}
\usepackage{fontawesome5} 
\usepackage[tikz]{bclogo}
\usepackage[framemethod=tikz]{mdframed}

\usepackage[export]{adjustbox}

\definecolor{rqblue}{HTML}{E8F1FA}
\definecolor{findinggreen}{HTML}{E8F9E8}
\definecolor{findingbrown}{HTML}{D4A37F}

\newtcolorbox{findingbox}[1][]{
    breakable,
    enhanced,
    sharp corners,
    boxrule=0pt,
    colback=findingbrown!20,
    colframe=findingbrown!20,
    frame hidden,
    borderline west={2pt}{0pt}{findingbrown},
    left=6pt,
    right=6pt,
    top=4pt,
    bottom=4pt,
    before skip=10pt,
    after skip=10pt,
    fontupper=\linespread{1.0}\selectfont,
    #1
}

\usepackage{lipsum}
\usepackage[page]{appendix}
\usepackage{setspace}

\renewcommand{\appendixtocname}{List of appendices}

\makeatletter
\let\oldappendix\appendices

\renewcommand{\appendices}{%
  \clearpage
  \renewcommand{\thesection}{\Roman{section}}
  \let\tf@toc\tf@app
  \addtocontents{app}{\protect\setcounter{tocdepth}{2}}
  \immediate\write\@auxout{%
    \string\let\string\tf@toc\string\tf@app^^J
  }
  \oldappendix
}%

\newcommand{\listofappendices}{%
  \begingroup
  \renewcommand{\contentsname}{\appendixtocname}
  \let\@oldstarttoc\@starttoc
  \def\@starttoc##1{\@oldstarttoc{app}}

  \renewcommand{\contentsname}{%
      \fontsize{15}{22}\selectfont\bfseries
      \appendixtocname
  }
  
  \begin{spacing}{1.5}  
  \tableofcontents
  \end{spacing}
  
  \endgroup
}

\makeatother

\title{\textsc{SubTokenTest}: A Practical Benchmark for Real-World Sub-token Understanding}

\author{Shuyang Hou$^*$ \quad Yi Hu$^*$ \quad Muhan Zhang$^\dagger$\\
  Institute for Artificial Intelligence, Peking University\\
  }

\begin{document}
\maketitle

\providecommand{\benchmarkname}{\textsc{SubTokenTest}} 

\begin{abstract}
    Recent advancements in large language models (LLMs) have significantly enhanced their reasoning capabilities. However, they continue to struggle with basic character-level tasks, such as counting letters in words---a problem rooted in their tokenization process. While existing benchmarks have highlighted this weakness through basic character operations, such failures are often dismissed due to lacking practical relevance. Yet, many real-world applications, such as navigating text-based maps or interpreting structured tables, rely heavily on precise sub-token understanding. In this regard, we introduce \benchmarkname{}, a comprehensive benchmark that assesses sub-token understanding through \textbf{practical}, \textbf{utility-driven} tasks. Our benchmark includes ten tasks across four domains and isolates tokenization-related failures by decoupling performance from complex reasoning. We provide a comprehensive evaluation of nine advanced LLMs. Additionally, we investigate the impact of test-time scaling on sub-token reasoning and explore how character-level information is encoded within the hidden states.

    \textbf{Code: } \href{https://github.com/Shu-Feather/subTokenTest.git}{\textcolor{black}{\faGithub} \texttt{Shu-Feather/SubTokenTest}}

    \textbf{Dataset: } \href{https://huggingface.co/datasets/Rain-air/SubTokenTest-QA}{\includegraphics[height=0.9em]{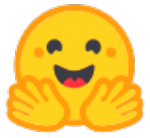} \texttt{Rain-air/SubTokenTest-QA}}


\end{abstract}
\renewcommand{\thefootnote}{\fnsymbol{footnote}} 
\footnotetext[1]{Equal contribution (order determined by coin flip).}
\footnotetext[2]{Corresponding author. \\Correspondence: \texttt{\{2200017797,huyi2002\}@stu.pku.edu.cn; muhan@pku.edu.cn}}

\section{Introduction}
Recent years have seen remarkable advancements in the reasoning capabilities of large language models (LLMs)~\citep{qwen2025qwen25technicalreport,deepseek_v3,gpt4,gpt4o}. Especially, reinforcement learning (RL) has further incentivized complex reasoning, resulting in powerful large reasoning models~\citep{deepseek_r1,o1,gpt5}. Despite these advances, LLMs can still surprisingly fail at seemingly trivial problems involving sub-word understanding. A widely-discussed failure is that LLMs struggle to   correctly answer the question: How many ``r''s are there in ``strawberry''?


This issue primarily stems from the way LLMs tokenize text. Trained on discrete tokens rather than individual characters, LLMs lack direct access to the characters that make up these tokens. Instead, they must infer information about character structures through statistical patterns present in the training corpus. This limitation becomes apparent when LLMs are asked to perform tasks that require precise character-level comprehension.

Several benchmarks have been proposed to evaluate the sub-token understanding of LLMs~\citep{edman2024cute,shin2024large}. These benchmarks typically focus on basic character-level operations such as spelling, character counting, insertion, deletion, and replacement. These studies consistently show that while LLMs can handle spelling, they struggle with most character composition tasks, revealing a gap in their sub-token handling capabilities.

\begin{figure}[t]
    \centering
    \includegraphics[width=\linewidth]{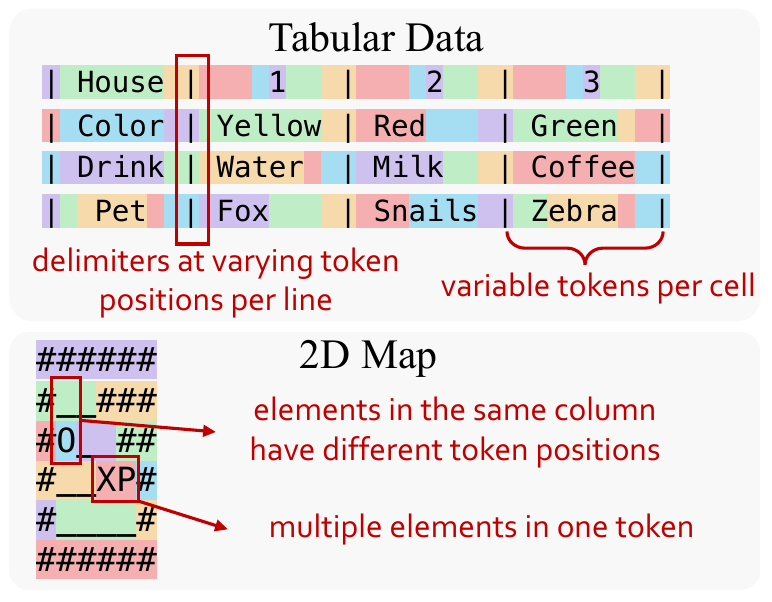}
    \caption{Example cases where tokenization may cause confusion in practical tasks.}
    \vspace{-10pt}
    \label{fig:map_table_tokenize}
\end{figure}

While these evaluations provide a foundational understanding of an LLM's capacity for character-level comprehension, the challenges they reveal are rarely encountered in everyday interactions with LLMs. For example, counting specific characters in a word is a trivial task for humans, and thus not commonly required from LLMs. Consequently, the failures of LLMs on these basic benchmarks have seemingly not received the attention commensurate with their practical implications.

In fact, many practical tasks, and even tasks in popular benchmarks, demand a high level of sub-word understanding capability, which can be easily overlooked~\cite{suzgun2023challenging,wang2025ragen}. 
Tasks involving agentic reasoning~\citep{wang2025ragen} and tabular data~\citep{van2024tabular,fang2024tabular}, for instance, require precise sub-word understanding.
Consider agentic tasks like navigating a 2D map in a game such as Sokoban, where each character denotes a game element (e.g., player, obstacle, box, etc.). Due to tokenization, as shown in Figure~\ref{fig:map_table_tokenize} bottom, multiple game elements may be contained within a single token, making it difficult for the model to precisely locate their row- and column-wise coordinates to accurately assess the game state. 
Besides, in table understanding tasks, as shown in Figure~\ref{fig:map_table_tokenize} top, the gap between tokenization granularity and the granularity of table cells makes it challenging for the model to correctly associate a table element with its coordinate.
Crucially, in these tasks, errors stemming from sub-token understanding are often \textbf{confounded} with reasoning errors, making it difficult to pinpoint whether the model’s failure is due to inadequate reasoning or weak sub-token comprehension.

\begin{figure}
    \centering
    \includegraphics[width=0.75\linewidth]{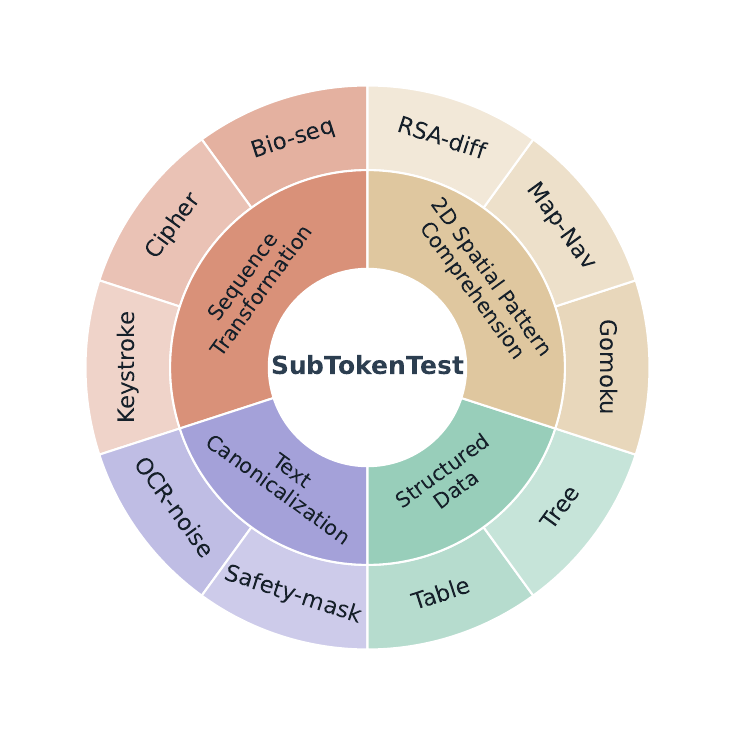}
    \caption{\benchmarkname{} categories.}
    \label{fig:pie_chart}
\end{figure}

To address this gap, we propose \benchmarkname{}, a comprehensive benchmark designed to assess sub-token understanding through the lens of \textbf{real-world utility}. A key design principle of our benchmark is the decoupling of perceptual accuracy from complex reasoning. By minimizing logical complexity, we isolate the model's ability to faithfully parse and manipulate the underlying character stream within token sequences.
As illustrated in Figure~\ref{fig:pie_chart}, \benchmarkname{} comprises ten diverse tasks spanning four critical domains, including
\textbf{Sequence Transformation}, which assesses precise, character-by-character execution of deterministic mappings;
\textbf{Text Canonicalization}, which requires normalizing text based on context and rules;
\textbf{Structured Data}, which evaluates the comprehension and rendering of aligned tables and trees; and
\textbf{2D Spatial Pattern Recognition}, which tests the ability to reconstruct a two-dimensional topology from token sequences.



We conduct a detailed evaluation of nine advanced LLMs on \benchmarkname{}, spanning both reasoning and non-reasoning models. We observe that while large-scale reasoning models can mitigate sub-token blindness to some extent, they do so at an extreme token cost. In contrast, smaller models exhibit notably poor performance. For reasoning models, we further test their test-time scaling and observe a inverted U-shaped performance curve in sub-token understanding as reasoning tokens increase. Moreover, through interpretability analyses, we examine whether internal representations encode character-level information across diverse text forms, including normal words, typos, random character sequences, and special symbols.

\section{Related Work}

\paragraph{Benchmarks for Sub-token Understanding.}
Existing research has primarily assessed LLMs' character-level comprehension through basic tasks. Early studies highlighted that LLMs lag behind humans in elementary operations like word length estimation and constrained text generation~\citep{efrat2023lmentry}. Later work expands to tasks such as insertion, deletion, replacement, reordering, and counting~\citep{edman2024cute, shin2024large, sims2025stochastok, uzan2025charbench}. While LLMs struggle with these tasks, which are trivial for humans, their failures in basic operations are rarely encountered in real-world interactions.
However, the consequences of this ``sub-token blindness'' extend far beyond simple character operation errors and have been shown to directly impede more complex, functional reasoning. For example, tokenization granularity affects numerical accuracy, with multi-digit tokenization degrading arithmetic performance as the model cannot distinguish individual digits~\citep{yang2024number, singh2024tokenization}. Despite these findings, the broader impact of sub-token understanding in practical tasks remains under-explored.

\paragraph{Mitigation Strategies for Tokenization Limitations.}
The limitations of Byte-Pair Encoding (BPE; \citealp{sennrich2016bpe}) have led to diverse architectural and prompting solutions. Stochastic tokenization methods like BPE-Dropout \citep{provilkov2020bpe} and StochasTok \citep{sims2025stochastok} introduce randomness during encoding to expose the model to various subword segmentations, thereby improving fine-grained understanding. Prompting strategies, such as ``divide-and-conquer'' techniques, where models spell out words before manipulation, have proven effective in character manipulation~\citep{xiong2025enhancing} and numerical tasks~\citep{yang2024number, hu2024case, hu2025beyond}. Recent architectural innovations \citep{pagnoni2025byte, yu2025scaling} suggest dynamically sized byte patches instead of fixed tokenization. However, even these token-free models still struggle with character-level tasks, highlighting the challenge of achieving robust sub-token understanding.

\paragraph{Mechanisms of Sub-token Understanding.}
Mechanistic studies investigate whether and to what extent LLMs internally encode character-level information. \citet{itzhak2022models, alajrami2023understanding, hiraoka2025spelling} show that models implicitly encode the composition of tokens within their representations without explicit supervision. \citet{kaplan2024tokens, wang2025word} further identifies an intrinsic ``detokenization'' process, where sub-word sequences are progressively reconstructed into coherent whole-word representations in the deeper layers. 

\section{Real-World Subtoken Evaluation}
\label{sec:real_world_subtoken}

To bridge the gap between rudimentary character-level operation evaluation and actual model utility, we propose \benchmarkname{}. Unlike existing benchmarks that focus on isolated atomic operations, \benchmarkname{} targets the real-world scenarios where sub-token blindness directly leads to functional failure. As illustrated in Figure~\ref{fig:pie_chart}, the benchmark is organized into four  domains:
\begin{enumerate}[
  topsep=0pt,      
  partopsep=0pt,   
  itemsep=0pt,     
  parsep=0pt,      
  leftmargin=*,    
]
\item \textbf{Sequence Transformation} evaluates the model's precision in executing deterministic, character-level mappings, including \textit{keystroke-level text editing}, \textit{cipher \& decipher}, and \textit{biological sequence manipulation}.
\item \textbf{Text Canonicalization} requires the model to normalize text based on both context and given rules. This includes \textit{OCR-noise canonicalization}, testing robustness to character-level perturbations, and \textit{safety-style masking}, demanding precise edition for sensitive data masking.
\item \textbf{Structured Data} evaluates the model’s ability to comprehend and render specific structures including \textit{aligned tables} and \textit{trees} through token streams, where token boundaries often misalign with data delimiters, as shown in Figure~\ref{fig:map_table_tokenize}.
\item \textbf{2D Spatial Pattern Recognition} focuses on the model’s ability to reconstruct a 2D topology from a 1D token sequence, through tasks including \textit{map navigation}, \textit{Gomoku state reading}, or \textit{RSA difference identification}.
\end{enumerate}

\subsection{Task Suite}
\label{sec:task_suite}
Below we provide a comprehensive overview of the data curation, core challenges, and evaluation metrics for each of the ten tasks in \benchmarkname{}. To establish a clear connection between practical challenges and their root causes, we map each high-level task to the underlying atomic character-processing abilities, summarized in Table~\ref{tab:task_ability_mapping}. These abilities include:
\textbf{Spelling}: mapping a token to its containing characters; 
\textbf{Inverse Spelling}: reconstructing a token from a sequence of characters; 
\textbf{Character Counting}: counting specific characterss within a token; 
\textbf{Character Indexing}: locating the exact position of a character within a token; 
\textbf{Character Substitution}: replacing a character in a token with another;
\textbf{Token Length}: determine the total number of characters in a token.
We provide example input-output traces for each task in Appendix~\ref{app:tasks}. 

\begin{table}[t]
\centering
\caption{Mapping between tasks in \benchmarkname{} and required atomic sub-token abilities.}
\label{tab:task_ability_mapping}
\resizebox{\linewidth}{!}{%
\begin{tabular}{cl}
\toprule
\textbf{\benchmarkname{} Task} & \textbf{Required Atomic Abilities} \\
\midrule
\textbf{Typewriter Effect Task} & Spelling \\
\textbf{Backspace Handling Task} & Inverse Spelling \\\midrule
\textbf{Cipher} & Spelling, Char Substitution \\
\textbf{Decipher} & Inverse Spelling, Char Substitution \\\midrule
\textbf{Biological Sequence} & Char Substitution \\\midrule
\textbf{OCR-noise Canonicalization} & Inverse Spelling, Char Substitution \\\midrule
\textbf{Safety-style Masking} & Char Indexing, Char Substitution, Token Length \\\midrule
\textbf{Aligned Tables} & Char Counting, Char Indexing, Token Length \\\midrule
\textbf{Tree Understanding} & Char Indexing, Token Length \\\midrule
\textbf{Map Navigation} & Char Counting, Char Indexing, Token Length \\\midrule
\textbf{Gomoku} & Char Counting, Char Indexing, Token Length \\\midrule
\textbf{RSA Difference} & Char Indexing, Token Length \\
\bottomrule
\end{tabular}
}
\end{table}

\paragraph{Evaluation Metrics.} We primarily rely on two key metrics for evaluation. \textbf{Exact match (EM)} measures strict correctness, computed as the proportion of predictions identical to the ground truth:
\(
\mathrm{EM} = \mathbb{I}\{p_i = g_i\}.
\),
where  $p_i$ is the extracted model prediction and $g_i$ is the ground-truth answer.
To provide a more granular assessment of near-correct responses, we calculate the \textbf{normalized Levenshtein similarity}.:
\(
\mathrm{Sim}(p_i,g_i) = 1 - \frac{d_{\mathrm{lev}}(p_i,g_i)}{\max(|p_i|,|g_i|)}
\),
where \( |\cdot| \) denotes sequence length and \( d_{\mathrm{lev}} \) is the Levenshtein distance. For detailed generation pipeline and evaluation metrics of each task, please refer to Appendix~\ref{app:metrics}.

\paragraph{Keystroke-level text editing.}
This task includes two complementary sub-tasks that evaluate the model's precision in tracking character-level modifications.
The \textbf{Keystroke Encoding Task} requires the model to decompose a given token sequence into a character-level typing log, testing its \textit{spelling} and token-to-character mapping. Conversely, the \textbf{Keystroke Decoding Task} provides a sequence of keystroke-level operations involves reconstructing a final text sequence from keystroke operations, assessing the model's \textit{inverse spelling} ability. Performance is measured by EM.

\paragraph{Cipher \& decipher.}
This task evaluates deterministic character-to-symbol and symbol-to-character transformations using \textbf{Morse and Caesar ciphers} at multiple difficulty tiers regarding original context length.
In the \textbf{Cipher} sub-task, models map plain text to encoded strings, which requires precise decomposition of input tokens into constituent characters (\textit{spelling}), followed by \textit{character substitution} within a symbolic space. The \textbf{Decipher} sub-task inverts this process: models must reconstruct original texts by transforming symbols back into characters through \textit{character substitution}, then aggregate these characters into tokens to test \textit{inverse spelling}. Evaluation is based on EM and sequence similarity.

\paragraph{Biological sequence manipulation.}
This task tests models on biology transformations by four sub-tasks: DNA complementation, RNA complementation, and bidirectional conversion between single-letter and three-letter protein code representations. All operations require deterministic symbol-to-symbol mapping, isolating and testing the model's \textit{character substitution} capability without requiring complex biological reasoning. Sequences are synthetically generated with controlled lengths to create multiple difficulty tiers. Here we use EM and sequence similarity.

\paragraph{OCR-noise canonicalization.}
This task challenges models to recover canonical text from prompts corrupted by character-level OCR-like perturbations, such as leet-speak substitutions (e.g., \texttt{a}$\rightarrow$$\alpha$, \texttt{u}$\rightarrow$$\upsilon$). Prompts span three categories (harmful, jailbreak, benign) with varying text lengths. Successful canonicalization requires the model to first identify and rectify corrupted characters through \textit{character substitution}, then aggregate these corrected units into coherent, valid tokens via \textit{inverse spelling}. Performance is assessed using both EM and sequence similarity.

\paragraph{Safety-style masking.}
This task evaluates the model's ability to identify and mask sensitive numeric information within templated, realistic contexts. Difficulty varies with context length and sensitive item density. Successful masking requires model to navigate potentially fragmented token sequences using \textit{token length} awareness to traverse preceding tokens, employ \textit{character indexing} to pinpoint exact digit boundaries within tokens overlapping the sensitive span, and execute precise \textit{character substitution} to replace target digits with masking symbols. We focus on phone numbers, credit cards, and 18-digit ID cards. Performance is measured using exact-match accuracy and F1 score, with F1 computed at the number level as a binary classification (redacted or not).

\paragraph{Aligned tables.}
This task requires models to render structured data into aligned tables using \LaTeX, Markdown, or ASCII formats under strict stylistic constraints. The core challenge is maintaining vertical alignment, as tokenizers often merge consecutive spaces or combine delimiters with adjacent cell content as shown in Figure~\ref{fig:map_table_tokenize}. 
To succeed, the model must master \textit{token length} calculation to ensure uniform column widths, \textit{character indexing} to place delimiters at precise boundaries, and \textit{character counting} to generate the exact number of required padding spaces. Performance is assessed via average content score and alignment rate detailed in Appendix~\ref{app:metrics}.

\paragraph{Tree understanding.}
This task evaluates model's ability to comprehend tree structures rendered in 2D text formats. Given a visual representation of a binary tree, the model is tested through two complementary subtasks: tree structure and tree path. To successfully understand the tree's topology, the model must utilize \textit{token length} to calculate vertical alignment and horizontal indentation, alongside \textit{character indexing} to identify the exact position of connectors (such as ``/'' and ``\textbackslash'') relative to node labels. Performance is measured mainly via EM for both structural and path queries, which is detailed in Appendix~\ref{app:metrics}.


\definecolor{SeqTrans}{HTML}{CB947D}    
\definecolor{TextCanon}{HTML}{A2A0D4}   
\definecolor{StructData}{HTML}{A3CCBB}  
\definecolor{Spatial2D}{HTML}{DAC7A4}   

\begin{table*}[th]
    \centering
    \caption{Performance on the \benchmarkname{} benchmark. \small Abbreviations: DS-V3/R1 (DeepSeek-V3/R1), Ins (Instruct), DS-Qwen (DeepSeek-distill-Qwen). Detailed sub-task results are provided in Appendix ~\ref{app:sub_task}.}
    \label{tab:benchmark_summary_final}
    \renewcommand{\arraystretch}{1.4}
    \resizebox{\textwidth}{!}{%
        \begin{tabular}{lll|cc|cc|cc|cccc}
            \toprule
            \multirow{2}{*}{\textbf{Group}} 
            & \multirow{2}{*}{\textbf{\makecell[l]{Benchmark \\ Task}}} 
            & \multirow{2}{*}{\textbf{Metric}} 
            & \multicolumn{2}{c|}{\textbf{DeepSeek}} 
            & \multicolumn{2}{c|}{\textbf{GPT}} 
            & \multicolumn{2}{c|}{\textbf{o}}
            & \multicolumn{4}{c}{\textbf{Qwen \& DS-distill-Qwen}} \\
            \cmidrule(lr){4-5} \cmidrule(lr){6-7} \cmidrule(lr){8-9} \cmidrule(l){10-13}
            & & 
            & \textbf{DS-V3} 
            & \textbf{DS-R1}
            & \textbf{GPT-4} 
            & \textbf{GPT-5} 
            & \textbf{\makecell{o4-mini\\(low)}} 
            & \textbf{\makecell{o4-mini\\(high)}}
            & \textbf{\makecell{Qwen-2.5\\-7B-Ins}} 
            & \textbf{\makecell{DS-Qwen-\\2.5-7B}} 
            & \textbf{\makecell{Qwen-2.5\\-32B-Ins}} 
            & \textbf{\makecell{DS-Qwen-\\2.5-32B}} \\
            \midrule

            \cellcolor{SeqTrans} & \multirow{2}{*}{\makecell[l]{\textbf{Keystroke}}}
            & 
            Encoding EM & 86.0\%  & 95.0\%  & 90.0\%  & \textbf{99.0}\%  & 95.0\%    & 96.0\%     & 2.0\%  & 0.0\%    & 13.0\%  & 42.0\%    \\
            \cellcolor{SeqTrans} & & 
            Decoding EM & 0.0\%   & 94.0\%  & 0.0\%   & \textbf{100.0}\% & 95.0\%    & 98.0\%     & 0.0\%  & 0.0\%    & 0.0\%   & 0.0\%\\
            \cellcolor{SeqTrans} & & 
            \cellcolor{lightgray}Token Num          & \cellcolor{lightgray}21         & \cellcolor{lightgray}2102       & \cellcolor{lightgray}22         & \cellcolor{lightgray}1011       & \cellcolor{lightgray}676    & \cellcolor{lightgray}1997    & \cellcolor{lightgray}1279 & \cellcolor{lightgray}1921   & \cellcolor{lightgray}1623  & \cellcolor{lightgray}2311    \\
            \hhline{~------------}
            \cellcolor{SeqTrans} & \multirow{2}{*}{\textbf{\makecell[l]{Cipher \\ \& Decipher}}}
            & Exact Match& 30.0\%  & \textbf{43.0}\%  & 25.0\%  & 40.0\%  & 14.0\%    & 38.0\%     & 0.0\%  & 0.0\%    & 0.0\%   & 0.0\%\\
            \cellcolor{SeqTrans} & & Similarity & 0.56    & \textbf{0.96}    & 0.50 & \textbf{0.96}    & 0.84      & 0.88       & 0.07   & 0.08& 0.09    & 0.12 \\
            \cellcolor{SeqTrans} & & \cellcolor{lightgray}Token Num          & \cellcolor{lightgray}814        & \cellcolor{lightgray}14971      & \cellcolor{lightgray}554        & \cellcolor{lightgray}10103      & \cellcolor{lightgray}3404   & \cellcolor{lightgray}24118   & \cellcolor{lightgray}25750& \cellcolor{lightgray}23094  & \cellcolor{lightgray}18645 & \cellcolor{lightgray}20711   \\
            \hhline{~------------}
            \cellcolor{SeqTrans} & \multirow{2}{*}{\textbf{\makecell[l]{Bio-seq}}}
            & Exact Match& 77.0\%  & 85.0\%  & 32.0\%  & 97.0\%  & 86.0\%    & \textbf{98.0}\%     & 0.0\%  & 0.0\%    & 0.0\%   & 2.0\%\\
            \cellcolor{SeqTrans} & & Similarity & 0.86    & 0.94    & 0.74    & \textbf{0.99}    & 0.76      & \textbf{0.99}       & 0.13   & 0.17& 0.29    & 0.37 \\
            \cellcolor{SeqTrans}\multirow{-9}{*}{\rotatebox[origin=c]{90}{\color{black}\textbf{\makecell{Sequence\\Transformation}}}} & & \cellcolor{lightgray}Token Num          & \cellcolor{lightgray}485        & \cellcolor{lightgray}3150       & \cellcolor{lightgray}55         & \cellcolor{lightgray}2358       & \cellcolor{lightgray}1447   & \cellcolor{lightgray}3768    & \cellcolor{lightgray}2338 & \cellcolor{lightgray}2247   & \cellcolor{lightgray}1555  & \cellcolor{lightgray}2773 \\
            \midrule

            \cellcolor{TextCanon} & \multirow{1}{*}{\textbf{\makecell[l]{OCR-noise}}}
            & Exact Match& 70.0\%  & 77.0\%  & 79.0\%  & \textbf{91.0}\%  & 20.0\%    & 47.0\%     & 0.0\%  & 0.0\%    & 12.0\%  & 1.0\%\\

            \cellcolor{TextCanon} & & Similarity & 0.92    & 0.93    & 0.95    & \textbf{0.99}    & 0.94      & 0.97       & 0.57   & 0.17& 0.94    & 0.88 \\
            
            \cellcolor{TextCanon} & & \cellcolor{lightgray}Token Num          & \cellcolor{lightgray}1366       & \cellcolor{lightgray}10151      & \cellcolor{lightgray}328        & \cellcolor{lightgray}7694       & \cellcolor{lightgray}619    & \cellcolor{lightgray}3325    & \cellcolor{lightgray}12614& \cellcolor{lightgray}13376  & \cellcolor{lightgray}12672 & \cellcolor{lightgray}3946 \\
            \hhline{~------------}
            \cellcolor{TextCanon} & \multirow{1}{*}{\textbf{\makecell[l]{Safety-mask}}}
            & Exact Match& 54.0\%  & 96.0\%  & 59.0\%  & \textbf{98.0}\%  & 86.0\%    & 88.0\%     & 0.0\%  & 0.0\%    & 8.0\%   & 2.0\%\\

            \cellcolor{TextCanon} & & F1         & 0.93  & \textbf{1.00}  & 0.92   & \textbf{1.00}  & 0.98     & 0.98     & 0.12 & 0.13   & 0.76  & 0.51    \\
            
            \cellcolor{TextCanon}\multirow{-5}{*}{\rotatebox[origin=c]{90}{\color{black}\textbf{\makecell{Text\\Canonicalization}}}} & & \cellcolor{lightgray}\cellcolor{lightgray}Token Num          & \cellcolor{lightgray}241        & \cellcolor{lightgray}3216       & \cellcolor{lightgray}239        & \cellcolor{lightgray}2579       & \cellcolor{lightgray}895    & \cellcolor{lightgray}3563    & \cellcolor{lightgray}514  & \cellcolor{lightgray}432    & \cellcolor{lightgray}289   & \cellcolor{lightgray}317 \\
            \midrule

            \cellcolor{StructData} & \multirow{2}{*}{\makecell[l]{\textbf{Table}}}
            & Content Score         & 0.69    & 0.78    & \textbf{0.95}    & 0.66    & 0.34      & 0.69       & 0.52   & 0.21& 0.55    & 0.44 \\
            \cellcolor{StructData} & & Alignment Rate        & 41.0\%  & 23.9\%  & 21.0\%  & 36.0\%  & 34.0\%    & \textbf{59.0\%}     & 0.0\%  & 57.1\%   & 6.0\%   & 1.2\%\\
            \cellcolor{StructData} & & \cellcolor{lightgray}Token Num          & \cellcolor{lightgray}425        & \cellcolor{lightgray}15281      & \cellcolor{lightgray}418        & \cellcolor{lightgray}9981       & \cellcolor{lightgray}2436   & \cellcolor{lightgray}23276   & \cellcolor{lightgray}13701& \cellcolor{lightgray}11705  & \cellcolor{lightgray}4874  & \cellcolor{lightgray}6990  \\
            \hhline{~------------}
            \cellcolor{StructData} & \multirow{2}{*}{\makecell[l]{\textbf{Tree}}}
            & Structure EM          & 78.0\%  & 95.0\%  & 43.0\%  & \textbf{96.0\%}  & 83.0\%    & 95.0\%     & 14.0\% & 11.0\%   & 25.0\%  & 32.0\%    \\
            \cellcolor{StructData} & & Path EM    & 93.0\%  & \textbf{99.0\%}  & 27.0\%  & 97.0\%  & 72.0\%    & 90.0\%     & 0.0\%  & 2.0\%    & 31.0\%  & 31.0\%    \\
            \cellcolor{StructData}\multirow{-6}{*}{\rotatebox[origin=c]{90}{\color{black}\textbf{\makecell{Structured\\Data}}}} & & \cellcolor{lightgray}Token Num          & \cellcolor{lightgray}446        & \cellcolor{lightgray}4038       & \cellcolor{lightgray}90         & \cellcolor{lightgray}1409       & \cellcolor{lightgray}1367   & \cellcolor{lightgray}2959    & \cellcolor{lightgray}3179 & \cellcolor{lightgray}5743   & \cellcolor{lightgray}3526  & \cellcolor{lightgray}5296 \\
            \midrule

            
            \cellcolor{Spatial2D} & \multirow{1}{*}{\textbf{\makecell[l]{Map-Nav \\ (Sokoban)}}}
            & Exact Match& 79.0\%  & \textbf{99.0\%}  & 48.0\%  & 98.0\%  & 74.0\%    & 89.0\%     & 29.0\% & 11.0\%   & 37.0\%  & 69.0\%    \\
            \cellcolor{Spatial2D} & & \cellcolor{lightgray}Token Num          & \cellcolor{lightgray}331        & \cellcolor{lightgray}1478       & \cellcolor{lightgray}44         & \cellcolor{lightgray}891        & \cellcolor{lightgray}1076   & \cellcolor{lightgray}1423    & \cellcolor{lightgray}41   & \cellcolor{lightgray}2839   & \cellcolor{lightgray}119   & \cellcolor{lightgray}1114 \\
            \hhline{~------------}
            \cellcolor{Spatial2D} & \multirow{1}{*}{\textbf{\makecell[l]{Map-Nav \\ (FrozenLake)}}}
            & Exact Match& 93.0\%  & \textbf{99.0\%}  & 59.0\%  & \textbf{99.0\%}  & 79.0\%    & 87.0\%     & 33.0\% & 18.0\%   & 47.0\%  & 80.0\%    \\
            \cellcolor{Spatial2D} & & \cellcolor{lightgray}Token Num          & \cellcolor{lightgray}289        & \cellcolor{lightgray}1010       & \cellcolor{lightgray}23         & \cellcolor{lightgray}602        & \cellcolor{lightgray}751    & \cellcolor{lightgray}889& \cellcolor{lightgray}21   & \cellcolor{lightgray}1918   & \cellcolor{lightgray}96    & \cellcolor{lightgray}1052 \\
            \hhline{~------------}
            \cellcolor{Spatial2D} & \multirow{2}{*}{\makecell[l]{\textbf{Gomoku}}}
            & Linear EM  & 53.0\%  & 93.0\%  & 36.0\%  & \textbf{97.0\%}  & 77.0\%    & \textbf{97.0\%}     & 32.0\% & 30.0\%   & 43.0\%  & 41.0\%    \\
            \cellcolor{Spatial2D} & &  Diagonal EM & 30.0\%  & 81.0\%  & 34.0\%  & 94.0\%  & 42.0\%    & \textbf{97.0\%}     & 30.0\% & 16.0\%   & 35.0\%  & 27.0\%    \\
            \cellcolor{Spatial2D} & &  \cellcolor{lightgray}Token Num          & \cellcolor{lightgray}2213       & \cellcolor{lightgray}13801      & \cellcolor{lightgray}472        & \cellcolor{lightgray}11618      & \cellcolor{lightgray}2069   & \cellcolor{lightgray}13037   & \cellcolor{lightgray}577  & \cellcolor{lightgray}7973   & \cellcolor{lightgray}859   & \cellcolor{lightgray}8423    \\
            \hhline{~------------}
            \cellcolor{Spatial2D} & \multirow{1}{*}{\textbf{\makecell[l]{RSA-diff}}}
            & F1         & 0.07  & 0.97  & 0.04  & \textbf{0.98}  & 0.58     & 0.95     & 0.01 & 0.00    & 0.01  & 0.00    \\
            \cellcolor{Spatial2D}\multirow{-9}{*}{\rotatebox[origin=c]{90}{\color{black}\textbf{\makecell{2D Spatial Pattern\\Comprehension}}}} & & \cellcolor{lightgray}Token Num          & \cellcolor{lightgray}1331       & \cellcolor{lightgray}12665      & \cellcolor{lightgray}79         & \cellcolor{lightgray}6624       & \cellcolor{lightgray}2744   & \cellcolor{lightgray}7237    & \cellcolor{lightgray}494  & \cellcolor{lightgray}1496   & \cellcolor{lightgray}99    & \cellcolor{lightgray}1251    \\
            
            \bottomrule
        \end{tabular}
    }
\end{table*}

\paragraph{Map navigation.}

This task requires models to comprehend 2D grid map representations which are frequently used in agentic tasks such as Sokoban and FrozenLake~\citep{wang2025ragen}, where characters represent game objects. However, as shown in Figure~\ref{fig:map_table_tokenize}, tokenization often merges these objects with adjacent spaces or neighboring elements, obscuring their coordinates. This task comprises five task primitives detailed in Appendix~\ref{app:tasks}.
To solve these tasks, the model has to utilize \textit{character indexing} to pinpoint each object's offset within a merged token, \textit{token length} to calculate its precise horizontal and vertical position in the grid, and \textit{character counting} to aggregate global statistics about object distributions. Performance is measured via EM accuracy.

\paragraph{Gomoku state reading.}
This task evaluates the model’s ability to interpret 2D Gomoku game states represented in ASCII. The model must detect multi-directional lines (horizontal, vertical, diagonal) to determine the game outcome.
Model has to utilize \textit{character indexing} to locate stones within merged tokens, apply \textit{token length} analysis to reconstruct the 2D grid structure, and execute directional \textit{character counting} to trace and verify five consecutive stones of the same color. Performance is measured via EM across Linear (horizontal/vertical) and Diagonal subsets.

\paragraph{RSA difference.}
This task evaluates the model's ability to compare two RSA fingerprint ``randomart'' patterns at the character level. Given a pair of ASCII grids, the model must identify all differing positions and report both their coordinates and character changes.
Models are required to perform pixel-like comparison across two 2D grids while maintaining positional accuracy: models must apply \textit{token length} counting and \textit{character indexing} to locate and align corresponding characters between the two patterns despite tokenization boundaries. Performance is assessed using the F1 score, calculated based on the coordinate-level match.

\section{Results}

\paragraph{Experiment Setup.} We evaluate diverse advanced LLMs. For API-based evaluations, we include GPT-4~\citep{gpt4}, GPT-5~\citep{gpt5}, o4-mini~\citep{o4-mini}, DeepSeek-V3~\citep{deepseek_v3}, and DeepSeek-R1~\citep{deepseek_r1}. DeepSeek-distill-Qwen~\citep{deepseek_r1} and Qwen-2.5-Instruct~\citep{qwen} are evaluated locally. Detailed configurations are provided in Appendix~\ref{app:model_setup}.
\benchmarkname{} features an automated task generation pipeline with controllable difficulty levels for each task, as detailed in Appendix~\ref{app:task_gen}. For the evaluation results presented in Table~\ref{tab:benchmark_summary_final}, we curate a test set composed of total 1,700 test instances, and the detailed composition is also provided in Appendix~\ref{app:test_setup}.





\paragraph{Main Results.}

\begin{figure*}[thbp]
    \centering
    
    \includegraphics[width=\textwidth]{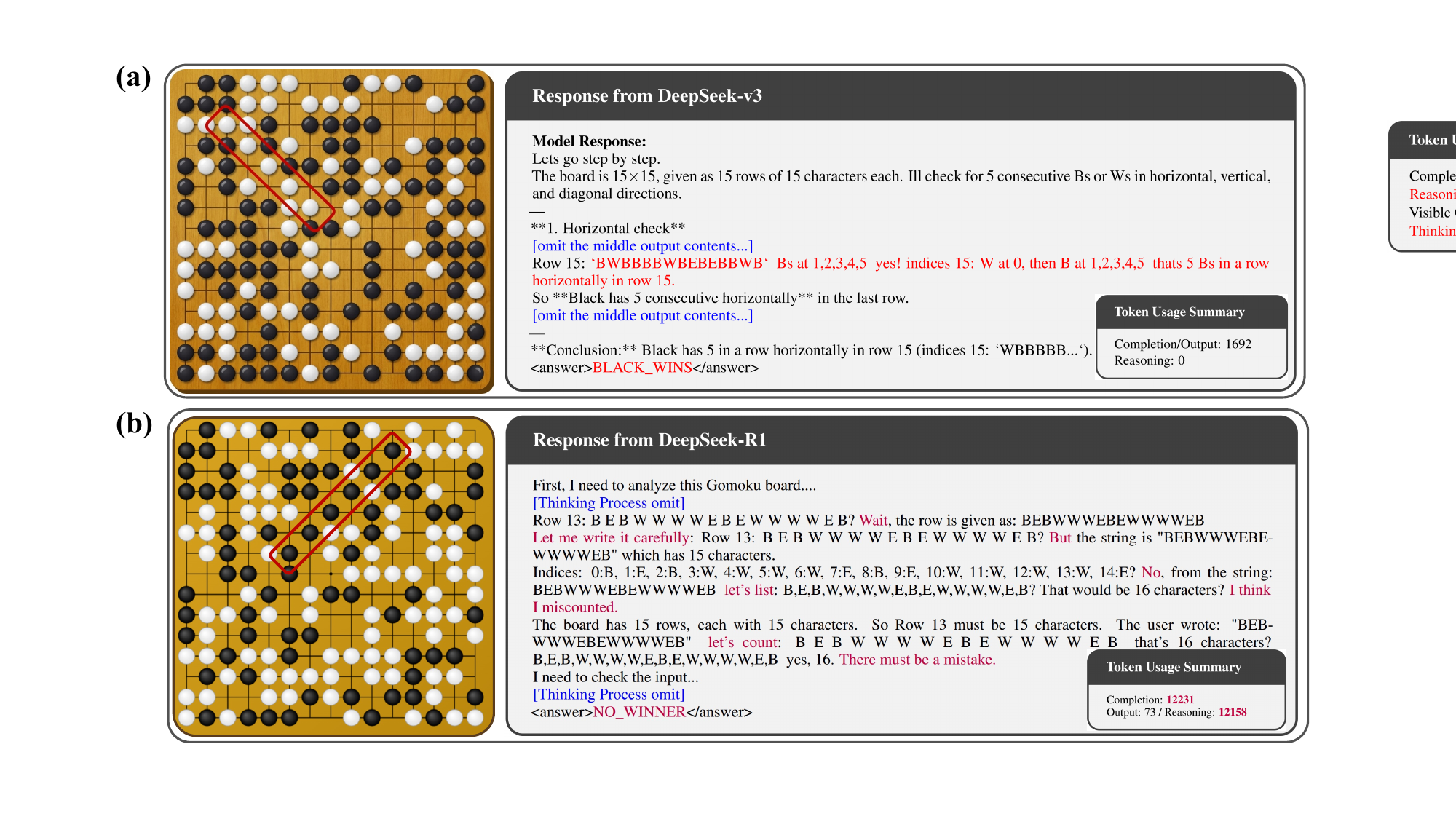}
    
    \caption{The two subfigures illustrate distinct error patterns: (a) tokenization-induced errors in DeepSeek-V3 (highlighted in \textcolor{red}{red}), and (b) overthinking errors in DeepSeek-R1 (highlighted in \textcolor{purple}{purple}).}
    \label{fig:gomoku}
\end{figure*}

Our full evaluation results are shown in Table~\ref{tab:benchmark_summary_final}. For large-scale models, reasoning models consistently outperform their non-reasoning counterparts. Comparing DeepSeek-R1 to DeepSeek-V3, and GPT-4 to GPT-5, we observe that \textbf{reasoning models successfully mitigate sub-token blindness} by explicitly decomposing tokens into constituent characters within their thinking traces. For instance, as illustrated in Figure~\ref{fig:gomoku}, DeepSeek-R1 decomposes each token character-by-character in the Gomoku task.
However, \textbf{this performance gain comes at a significant computational cost}. Reasoning models frequently consume tens or even hundreds of times more tokens than their base models to achieve these results. Furthermore, this capability is highly sensitive to the allocated reasoning budget. Our testing of the o4-mini series shows that reducing the reasoning budget from ``high'' to ``low'' causes a sharp accuracy drop, suggesting that even specialized reasoning models revert to sub-token-blind behavior when inference tokens are constrained. More detailed results are provided in Appendix~\ref{app:reasoning_budget}, and we explore the relationship between token budget and performance further in Section~\ref{sec:ttbc}.

Across most tasks, we observe a significant performance gap based on model scale: \textbf{large-scale models consistently outperform smaller models by a wide margin}. However, \textbf{within the smaller parameter range} (7B to 32B), a surprising trend emerges: \textbf{reasoning models do not always outperform their non-reasoning counterparts}. In fact, DeepSeek-R1-Distill-Qwen-7B consistently lags behind its non-reasoning base model across the majority of our benchmarks.
These smaller models remain consistently confused by tokenization-induced errors. In such cases, the extended CoT often adds noise instead of correcting errors, as the model lacks the internal representation capacity to effectively track character-level states.




\paragraph{Error analysis.}
To understand why models fail, we categorize the observed error patterns into two primary error categories: tokenization-induced misinterpretations and overthinking. Detailed case studies are provided in Appendix \ref{app:benchmark_error}.

\textbf{Tokenization-induced errors} occur when the model's processing granularity misaligns with the required character-level precision, leading to fundamental misinterpretations. This issue appears across all tasks in \benchmarkname{}. For instance, in the Gomoku task, as shown in Figure~\ref{fig:gomoku} (a), the model fails to correctly parse the board states because multiple positional symbols are often merged into a single token, preventing accurate extraction of individual board positions and corrupting the subsequent game state analysis.

The second pattern is \textbf{overthinking}, particularly prevalent in reasoning models. This occurs when models generate overly long and redundant CoTs for intuitive tasks, consuming a large number of tokens without improving, and sometimes even harming accuracy. For example, in Figure~\ref{fig:gomoku} (b), DeepSeek-R1 are trapped in recursive reasoning cycles just to decompose a simple row of board states, consuming over 12k tokens. 





\subsection{How Does Test-Time Scaling Affects Sub-token Understanding?}
\label{sec:ttbc}
\begin{figure}[tb]
    \centering
    \includegraphics[width=\linewidth]{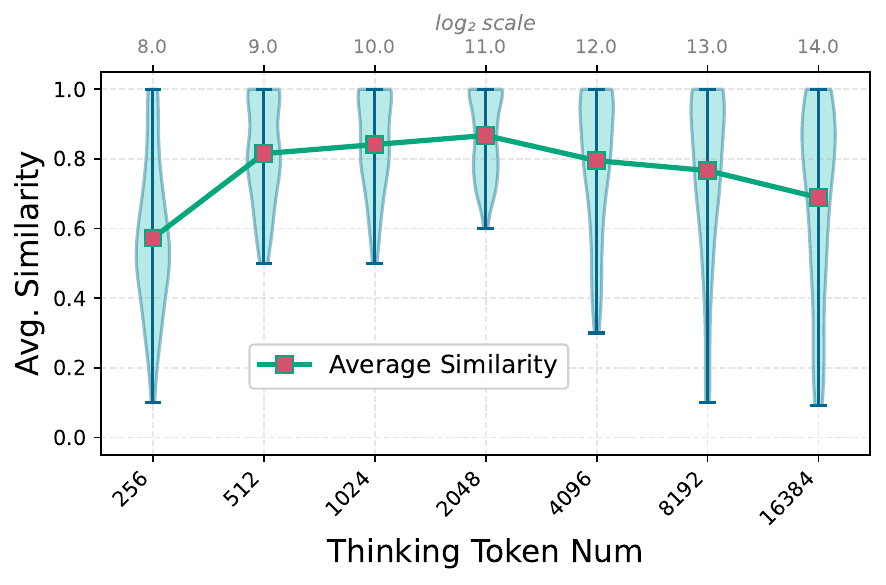}
    \caption{The effects of the number of thinking tokens on the task performance. Here we evaluate DS-distill-Qwen-2.5-7B on \textit{Biological Sequence Manipulation} with the metric of normalized similarity.}
    \label{fig:tts_violin}
\end{figure}

Reasoning models have shown impressive capabilities across complex tasks, including \benchmarkname{}. This success is largely attributed to test-time scaling, where models generate extended thinking traces to deliberate over difficult problems, as also reflected in the token usage metrics in Table~\ref{tab:benchmark_summary_final}. However, recent findings suggest that overly increasing reasoning length can trigger ``overthinking'', leading to performance degradation. Empirical studies across various benchmarks have shown an inverted U-shaped relationship between reasoning length and accuracy: performance improves initially but eventually declines as reasoning chains become overly redundant~\citep{tts1, tts2, tts3, tts4}.

To investigate this effect within the sub-token domain, we follow \citet{DBLP:journals/corr/abs-2506-04210} and implement a Test-Time Budget Control (TTBC) method. We explicitly modulate the length of the model's thinking trace by enforcing a strict token budget, $t_{\text{exact}}$. If a model attempts to terminate its reasoning prematurely, we inject a continuation cue (e.g., ``Wait'') to elicit further deliberation; conversely, traces exceeding the budget are truncated. We evaluate this on the \textit{Biological Sequence Manipulation} task using DeepSeek-R1-Distill-Qwen-7B, measuring performance via a length-normalized similarity score based on Levenshtein distance. The TTBC, task datasets, and evaluation methods are detailed in Appendix~\ref{app:ttbc}.

As shown in Figure~\ref{fig:tts_violin}, our results confirm the presence of the inverse U-shaped curve in sub-token tasks. Performance peaks at a budget of approximately 2048 tokens before suffering a significant decline at higher budgets. We identify three distinct phases in this scaling behavior: an increasing phase (256-512 tokens) where additional reasoning improves performance; a plateau phase (1024-2048 tokens) characterized by stable performance as the model conducts thorough verification; and finally a decreasing phase (exceeding 2048 tokens) where overthinking leads to redundant reasoning that degrades accuracy. Detailed error analysis are provided in Appendix~\ref{app:ttbc_error}.

\subsection{Do LLMs Encode Character-level Information in Hidden States?}
\label{sec:interp}
\begin{figure}[t]
    \centering
    \includegraphics[width=\linewidth]{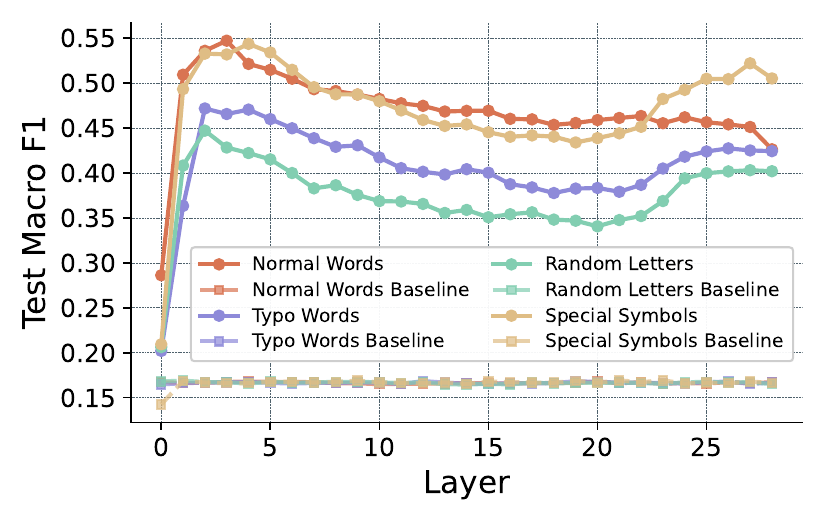}
    \caption{The Macro F1 results of linear probing the last token of certain token sequences. The dot line is the experimental group, and the square line is the corresponding baseline trained with shuffled labels.}
    \label{fig:linear_probe}
    \vspace{-5pt}
\end{figure}

We conduct an interpretability analysis to examine how LLMs encode character-level information in the hidden representations across various input formats. In \benchmarkname{}, we cover multiple text forms, including normal words, typo words (\textit{OCR-noise}), random letters (\textit{keystroke}) and special symbols (\textit{map-nav}, \textit{RSA-diff}, \textit{Gomoku}). We aim to probe the character-level information in each layer's hidden states given these various forms of texts. 

\paragraph{Probing Method.} We perform linear probing on the hidden representations of Qwen-2.5-7B-Instruct. Following the observation that the last token of a sequence typically aggregates information for preceding units \citep{kaplan2024tokens, wang2025word}, we extract the hidden states $h_\ell$ from the final token at each layer $\ell$ as our probing targets.
We frame the character-level awareness as a multi-character count prediction task. For a given input string, we define a dataset-specific alphabet $\mathcal{A}$ of size $|\mathcal{A}|$. The goal of the probe is to predict the ``bag-of-characters'' count vector $y = (y_1, y_2, \dots, y_{|\mathcal{A}|}) \in \mathbb{N}^{|\mathcal{A}|}$, where $y_m$ represents the frequency of character $a_m$ in the input. For each layer, we train a linear classifier to map $h_\ell$ to these counts, modeling the task as a $(K+1)$-way classification problem, where $K$ is the maximum count observed. The probes are optimized using cross-entropy loss. To ensure the probes reflect actual representation rather than label memorization, we compare performance against a baseline trained on shuffled labels.

To evaluate the layer-wise ``decodability'' of this information, we use the Macro-averaged F1 score. We first calculate the F1 score for each character individually by averaging across all possible count classes, and then take the uniform average across the entire alphabet. More training and evaluation details are provided in Appendix~\ref{app:linear_probe}.

\paragraph{Results.}
As illustrated in Figure~\ref{fig:linear_probe}, the shuffled baselines maintain F1 scores around 0.16 across all four word types, which is substantially lower than the performance of the normally trained probes, confirming the effectiveness of the linear probing method. Across all sequence types, we observe a consistent pattern in how character-level information evolves across layers. In the embedding layer, the F1 score is predictably low, as the last token has not yet integrated information from the preceding tokens. However, character awareness of the whole word sequence surges significantly within the first 2–3 layers, suggesting a rapid internal reconstruction process.
After this initial peak, awareness slightly plateaus or declines until approximately the 20th layer. For typo words, special symbols, and random letters, we observe a secondary rise in F1 scores in these deeper layers, whereas the performance for normal words remains stable. Notably, the model's internal representations consistently retain more information for normal words and special symbols than for typo words, which in turn outperform random letters. This hierarchy suggests that the model’s character-level ``vision'' is heavily influenced by the text forms, which partially explains why models struggle more with non-semantic (\textit{keystroke decoding}) or perturbed token (\textit{OCR-noise}) tasks.

\section{Conclusions}

We introduce \benchmarkname{}, a comprehensive benchmark designed to assess sub-token understanding in LLMs through real-world tasks. Through comprehensive evaluation, we reveal that large-scale reasoning models mitigate sub-token errors at a high token cost, and are sensitive to reasoning budgets, while smaller models exhibit poor performance. Additionally, we identify an inverted U-shaped relationship between reasoning effort and task performance in sub-token tasks. Moreover, probing results reveal how character-level information is encoded across model layers, with sub-token awareness evolving differently depending on the input format.


\section*{Limitations}
While this work provides a comprehensive benchmark for assessing sub-token understanding in LLMs, it is important to note that we do not propose solutions for the challenges identified. Our goal is to evaluate the current state of LLMs' ability to handle sub-token information, leaving further improvements for future work.

Additionally, our interpretability analysis is limited to a linear probe that provides some intuitions into how models process sub-token information. However, this approach does not fully explain the complete circuits by which models handle sub-token data throughout the entire process. A deeper, more comprehensive analysis of these circuits remains an open direction for future research.

\section*{Ethical Considerations}

This work propose a new benchmark to test the sub-token understanding in LLMs. We acknowledge the potential ethical implications of developing benchmarks that evaluate the limits of LLMs, particularly with respect to their accuracy and biases in handling character-level information. Our aim is to enhance the transparency and reliability of these models, ensuring they can be more effectively applied in practical scenarios.

AI assistants were utilized for language polishing and refinement, strictly limited to improving the fluency and clarity the text. All technical content, experimental results, analyses, and conclusions remain the original work of the authors.

\bibliography{custom}


\newpage
\listofappendices

\begin{appendices}
\section{Task Descriptions and Examples}
\label{app:tasks}

SubTokenTest benchmark is a collection of independent tasks for large language models. This appendix provides basic descriptions and illustrative examples for each task in the SubTokenTest benchmark.

\subsection{Keystroke-level Text Editing}
\label{app:task10}

\subsubsection*{Task Description}

This benchmark simulates human typing dynamics through progressive character growth and backspace resolution, under different prompt styles and difficulty settings.

\subsubsection*{Task Examples}
\begin{figure*}
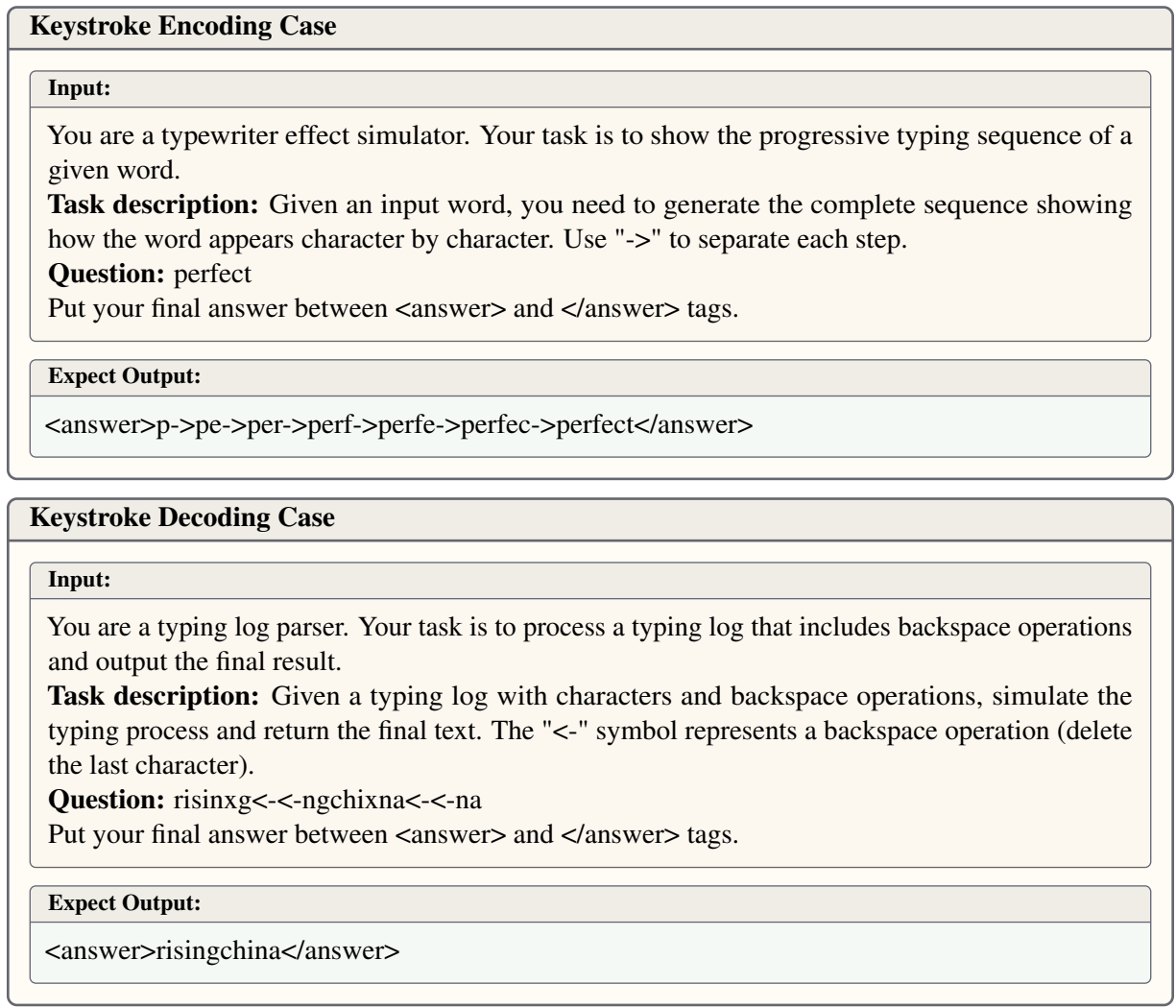

\begin{tcolorbox}[taskbox, title=Keystroke Encoding Case]
    \begin{tcolorbox}[examplebox={inputbg}{Input:}]
        You are a typewriter effect simulator. Your task is to show the progressive typing sequence of a given word.

    \textbf{Task description:} Given an input word, you need to generate the complete sequence showing how the word appears character by character. Use "->" to separate each step.
    
    \textbf{Question:} perfect

    Put your final answer between <answer> and </answer> tags.
    \end{tcolorbox}
    
    \begin{tcolorbox}[examplebox={outputbg}{Expect Output:}]
<answer>p->pe->per->perf->perfe->perfec->perfect</answer>
    \end{tcolorbox}

\end{tcolorbox}

\begin{tcolorbox}[taskbox, title=Keystroke Decoding Case]
    \begin{tcolorbox}[examplebox={inputbg}{Input:}]
        You are a typing log parser. Your task is to process a typing log that includes backspace operations and output the final result.

    \textbf{Task description:} Given a typing log with characters and backspace operations, simulate the typing process and return the final text. The "<-" symbol represents a backspace operation (delete the last character).
    
    \textbf{Question:} risinxg<-<-ngchixna<-<-na

    Put your final answer between <answer> and </answer> tags.
    \end{tcolorbox}
    
    \begin{tcolorbox}[examplebox={outputbg}{Expect Output:}]
<answer>risingchina</answer>
    \end{tcolorbox}

\end{tcolorbox}
\caption{Task examples of \textit{Keystroke-level Text Editing}.}
\label{fig:task_keystroke}
\end{figure*}

Figure~\ref{fig:task_keystroke} gives detailed task examples to both Keystroke encoding task and Keystroke decoding task.

\subsection{ Cipher \& Decipher}
\label{app:task4}

\subsubsection*{Task Description}

This suite probes model robustness on classical encoding and decoding tasks, including Morse code and Caesar ciphers.

\subsubsection*{Task Examples}
\begin{figure*}
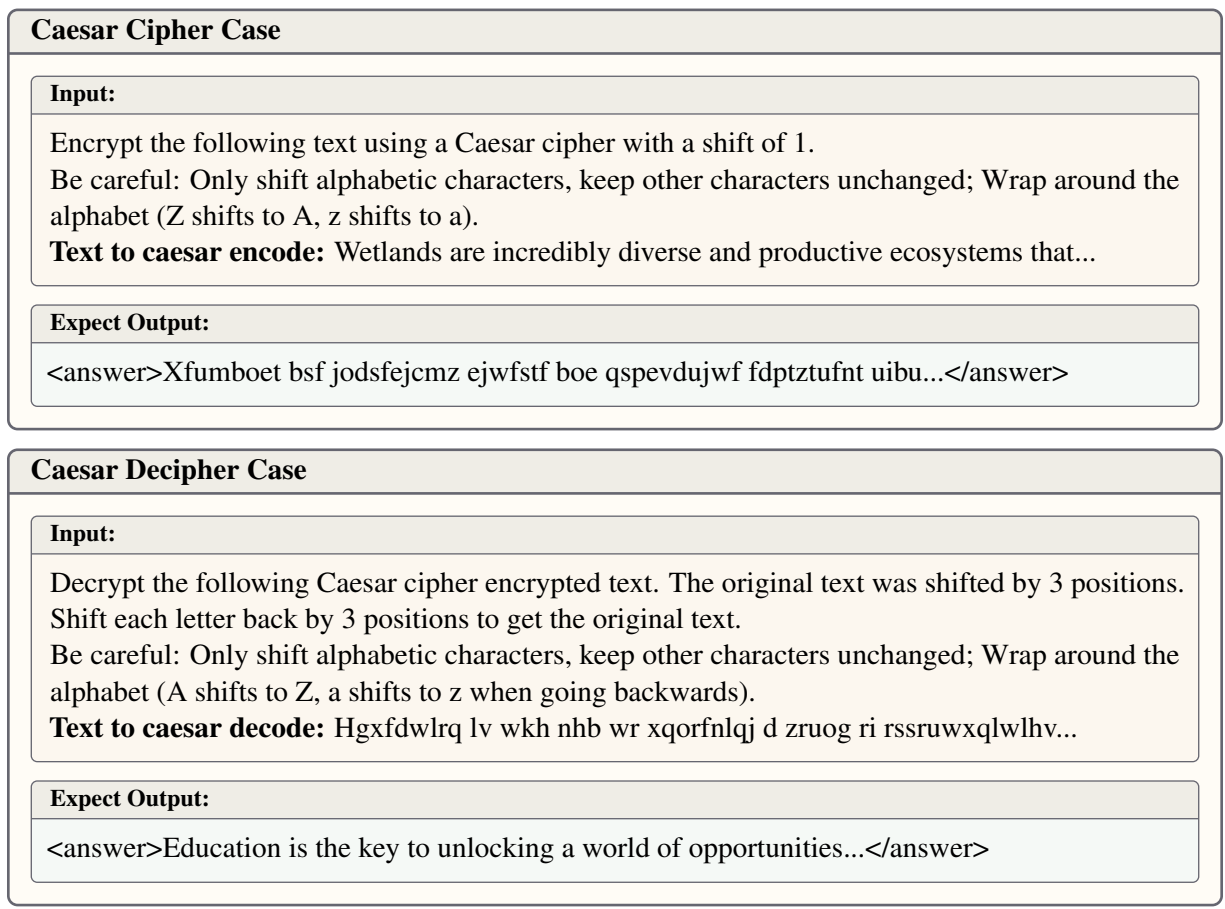


\begin{tcolorbox}[taskbox, title=Caesar Cipher Case]
    \begin{tcolorbox}[examplebox={inputbg}{Input:}]
        Encrypt the following text using a Caesar cipher with a shift of 1. 

Be careful: 
Only shift alphabetic characters, keep other characters unchanged; Wrap around the alphabet (Z shifts to A, z shifts to a).

        \textbf{Text to caesar encode:}
        Wetlands are incredibly diverse and productive ecosystems that...
    \end{tcolorbox}
    
    \begin{tcolorbox}[examplebox={outputbg}{Expect Output:}]
<answer>Xfumboet bsf jodsfejcmz ejwfstf boe qspevdujwf fdptztufnt uibu...</answer>
    \end{tcolorbox}

\end{tcolorbox}

\begin{tcolorbox}[taskbox, title=Caesar Decipher Case]
    \begin{tcolorbox}[examplebox={inputbg}{Input:}]
        Decrypt the following Caesar cipher encrypted text. The original text was shifted by 3 positions. Shift each letter back by 3 positions to get the original text.

Be careful: 
Only shift alphabetic characters, keep other characters unchanged; Wrap around the alphabet (A shifts to Z, a shifts to z when going backwards).

        \textbf{Text to caesar decode:}
        Hgxfdwlrq lv wkh nhb wr xqorfnlqj d zruog ri rssruwxqlwlhv...
    \end{tcolorbox}
    
    \begin{tcolorbox}[examplebox={outputbg}{Expect Output:}]
<answer>Education is the key to unlocking a world of opportunities...</answer>
    \end{tcolorbox}

\end{tcolorbox}
\caption{Task examples of Caesar \textit{Cipher \& Decipher}.}
\label{fig:task_cipher2}
\end{figure*}

Figure~\ref{fig:task_cipher1} and Figure~\ref{fig:task_cipher2} give detailed task examples to both Cipher \& Decipher tasks.
\begin{figure*}
\begin{tcolorbox}[taskbox, title=Morse Cipher Case]
    \begin{tcolorbox}[examplebox={inputbg}{Input:}]
        Convert the following text to Morse code. Use dots (.) and dashes (-) for the Morse code symbols, and separate each letter with a space. Use forward slash (/) for spaces between words.

        Be careful, the words between <answer> and </answer> tags should be your hand-in final answer.

        \textbf{Text to encode:}
        The Renaissance period
    \end{tcolorbox}
    
    \begin{tcolorbox}[examplebox={outputbg}{Expect Output:}]
        \begin{verbatim}
<answer>
- .... . / .-. . -. .- .. ... ... .- -. -.-. . / .--. . .-. .. --- -.. --..-- /
</answer>
        \end{verbatim}
    \end{tcolorbox}

\end{tcolorbox}

\begin{tcolorbox}[taskbox, title=Morse Decipher Case]
    \begin{tcolorbox}[examplebox={inputbg}{Input:}]
        Convert the following Morse code back to English text. The Morse code uses dots (.) and dashes (-), with spaces separating letters and forward slashes (/) representing spaces between words.
Be careful, as morse code DO NOT distinguish between uppercase and lowercase letters, your all answer words should be UPPERCASE.

        The words between <answer> and </answer> tags should be your hand-in final answer.

        \textbf{Text to decode:}
        \begin{verbatim}
- .... . / --. .-. . .- - / .-- .- .-.. .-.. / --- ..-. / -.-. .... .. -. .- /
        \end{verbatim}
    \end{tcolorbox}
    
    \begin{tcolorbox}[examplebox={outputbg}{Expect Output:}]
<answer>THE GREAT WALL OF CHINA</answer>
    \end{tcolorbox}

\end{tcolorbox}
\caption{Task examples of Morse \textit{Cipher \& Decipher}.}
\label{fig:task_cipher1}
\end{figure*}

\subsection{ Biological Sequence Manipulation}
\label{app:task3}

\subsubsection*{Task Description}

This benchmark evaluates molecular sequence reasoning capabilities, including DNA and RNA complementary strand generation and bidirectional protein translation between one-letter and three-letter amino acid codes, using explicitly provided pairing rules.

\subsubsection*{Task Examples}
\begin{figure*}
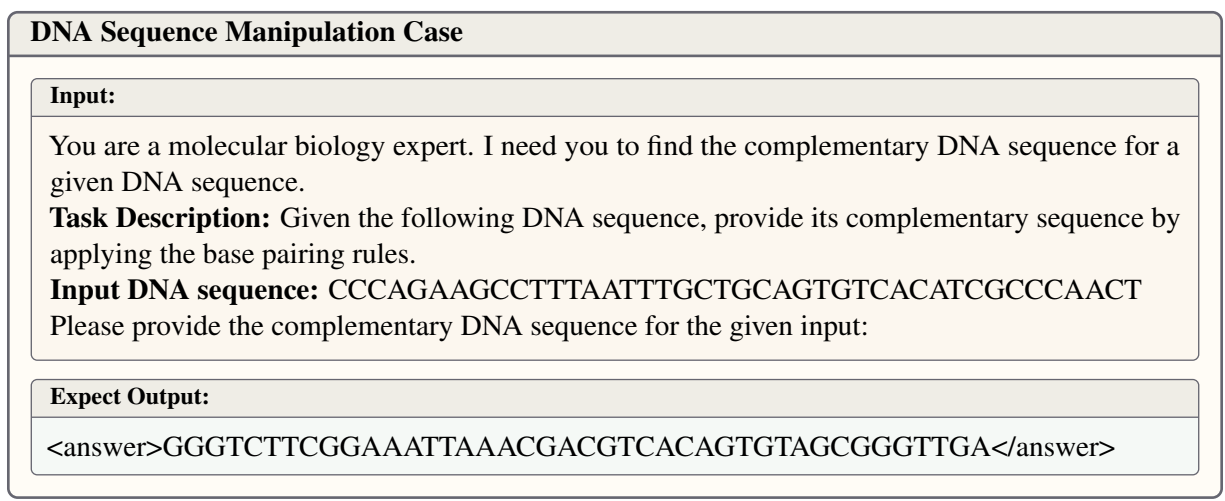

\begin{tcolorbox}[taskbox, title=DNA Sequence Manipulation Case]
    \begin{tcolorbox}[examplebox={inputbg}{Input:}]
        You are a molecular biology expert. I need you to find the complementary DNA sequence for a given DNA sequence.

        \textbf{Task Description:}
        Given the following DNA sequence, provide its complementary sequence by applying the base pairing rules.

        \textbf{Input DNA sequence:}
        CCCAGAAGCCTTTAATTTGCTGCAGTGTCACATCGCCCAACT
        
        Please provide the complementary DNA sequence for the given input:
    \end{tcolorbox}
    
    \begin{tcolorbox}[examplebox={outputbg}{Expect Output:}]
<answer>GGGTCTTCGGAAATTAAACGACGTCACAGTGTAGCGGGTTGA</answer>
    \end{tcolorbox}

\end{tcolorbox}
\caption{Task examples of DNA \textit{Biological Sequence Manipulation}.}
\label{fig:task_bio1}
\end{figure*}

Figure~\ref{fig:task_bio1}, Figure~\ref{fig:task_bio2} and Figure~\ref{fig:task_bio3} give detailed task examples to four types of biological sequence manipulation.

\begin{figure*}
\begin{tcolorbox}[taskbox, title=RNA Sequence Manipulation Case]
    \begin{tcolorbox}[examplebox={inputbg}{Input:}]
        You are a molecular biology expert. I need you to find the complementary RNA sequence for a given RNA sequence.

        \textbf{Task Description:}
        Given the following RNA sequence, provide its complementary sequence by applying the base pairing rules.

        \textbf{Input RNA sequence:}
        ACGUUCGAAGCCCUGCUUUCCAAAUCUGCUCAGACCCUA
        
        Please provide the complementary RNA sequence for the given input:
    \end{tcolorbox}
    
    \begin{tcolorbox}[examplebox={outputbg}{Expect Output:}]
<answer>UGCAAGCUUCGGGACGAAAGGUUUAGACGAGUCUGGGAU</answer>
    \end{tcolorbox}

\end{tcolorbox}
\caption{Task examples of RNA \textit{Biological Sequence Manipulation}.}
\label{fig:task_bio2}
\end{figure*}

\begin{figure*}
\begin{tcolorbox}[taskbox, title=Protein Sequence Manipulation Case]
    \begin{tcolorbox}[examplebox={inputbg}{Input:}]
        You are a biochemistry expert. I need you to convert a protein sequence from three-letter amino acid codes to one-letter codes.

        \textbf{Task Description:}
        Convert the following protein sequence from three-letter amino acid codes to one-letter codes.

        \textbf{Input protein sequence (three-letter codes):}
        PRO-MET-ASP-TRP-ARG-GLY-ALA-PRO
        
        Please convert the given protein sequence to one-letter amino acid codes:
    \end{tcolorbox}
    
    \begin{tcolorbox}[examplebox={outputbg}{Expect Output:}]
<answer>PMDWRGAP</answer>
    \end{tcolorbox}

\end{tcolorbox}

\begin{tcolorbox}[taskbox, title=Protein Sequence Manipulation Case]
    \begin{tcolorbox}[examplebox={inputbg}{Input:}]
        You are a biochemistry expert. I need you to convert a protein sequence from one-letter amino acid codes to three-letter codes.

        \textbf{Task Description:}
        Convert the following protein sequence from one-letter amino acid codes to three-letter codes.

        \textbf{Input protein sequence (one-letter codes):}
        IMVRHWPATPAVT
        
        Please convert the given protein sequence to three-letter amino acid codes:
    \end{tcolorbox}
    
    \begin{tcolorbox}[examplebox={outputbg}{Expect Output:}]
<answer>ILE-MET-VAL-ARG-HIS-TRP-PRO-ALA-THR-PRO-ALA-VAL-THR</answer>
    \end{tcolorbox}

\end{tcolorbox}
\caption{Task examples of Protein \textit{Biological Sequence Manipulation}.}
\label{fig:task_bio3}
\end{figure*}

\subsection{OCR-noise Canonicalization}
\label{app:task1}

\subsubsection*{Task Description}

This benchmark evaluates large language models on their ability to restore adversarially perturbed prompts to their canonical form. Perturbations include leetspeak substitutions (e.g., \texttt{a}$\rightarrow$$\alpha$, \texttt{u}$\rightarrow$$\upsilon$), character insertions (e.g., \texttt{sub\_to\_ken}), and mixed noise patterns. In this article, we focus on the leetspeak substitutions perturbations.

\subsubsection*{Task Example}
Figure~\ref{fig:task_ocr} gives a detailed task example to OCR-noise Canonicalization task.

\begin{figure*}
\begin{tcolorbox}[taskbox, title=OCR-noise Canonicalization Case]
    \begin{tcolorbox}[examplebox={inputbg}{Input:}]
        \textbf{System Prompt:}
        You are an expert text normalization system specializing in adversarial prompt detection and canonicalization.

        Your task is to:
        
        1. Analyze the given perturbed text that may contain character substitutions (like leet speak: '0' for 'o', '1' for 'i')
        
        2. Restore the text to its original, canonical form
        
        3. Place your final canonicalized answer between <answer> and </answer> tags

        \textbf{User Query:}
        
        Perturbed text:


        Please canonicalize this text and provide your answer between <answer> and </answer> tags.
        
    \end{tcolorbox}
    
    \vspace{3pt}
    
    \begin{tcolorbox}[examplebox={outputbg}{Expected Output:}]
        <answer>as an author working on...</answer>
    \end{tcolorbox}
\end{tcolorbox}
\caption{A task example of \textit{OCR-noise Canonicalization}.}
\label{fig:task_ocr}
\end{figure*}

\subsection{Safety-style Masking}
\label{app:task5}

\subsubsection*{Task Description}

This task assesses a model’s ability to detect and mask sensitive numerical information, such as phone numbers (e.g., \texttt{+1 89218772549} $\rightarrow$ \texttt{+1 892****2549}), identification numbers (e.g., \texttt{482403118361253800} $\rightarrow$ \texttt{482403**********00}), and credit card numbers (e.g., \texttt{6214 6972 3200 9219} $\rightarrow$ \texttt{621469******9219}), embedded within realistic textual contexts, while respecting context-specific reveal and masking rules.

\subsubsection*{Task Examples}
Figure~\ref{fig:task_mask} gives a detailed task example to Safety-style Masking task.

\begin{figure*}
\begin{tcolorbox}[taskbox, title=Safety-style Masking Case]
    \begin{tcolorbox}[examplebox={inputbg}{Input:}]
        You are a data privacy assistant. Your task is to redact (mask) sensitive information in the given text according to specific rules.

        \textbf{Text to redact:}
        Dear Sir/Madam, We would like to verify the phone number linked to your credit card. Please input this phone number in the space provided: +81 40516598483. Lastly, we would need the ID card number linked to your credit card, kindly input it in the field marked: 776244289659577527. Customer Service Team
    \end{tcolorbox}
    
    \begin{tcolorbox}[examplebox={outputbg}{Expect Output:}]
<answer>Dear Sir/Madam, We would like to verify the phone number linked to your credit card. Please input this phone number in the space provided: +81 405****8483. Lastly, we would need the ID card number linked to your credit card, kindly input it in the field marked: 776244**********27. Customer Service Team</answer>
    \end{tcolorbox}

\end{tcolorbox}
\caption{A task example of \textit{Safety-style Masking}.}
\label{fig:task_mask}
\end{figure*}

\subsection{ Aligned Table}
\label{app:task2}

\subsubsection*{Task Description}

This task tests whether models can accurately render structured descriptions into perfectly aligned tables in \LaTeX, Markdown, or plain text formats, while preserving both cell content and delimiter geometry.

\subsubsection*{Task Examples}
Figure~\ref{fig:task_table1} and Figure~\ref{fig:task_table2} give detailed task examples to four types of Aligned Table task.

\begin{figure*}
\begin{tcolorbox}[taskbox, title=Aligned Table Case]
    \begin{tcolorbox}[examplebox={inputbg}{Input:}]
        You are tasked with creating a well-formatted table based on the given information.

        \textbf{Task Description:}
        Given a description of table content, you need to output a properly formatted \{LATEX / MARKDOWN / TEXT\} table.

        \textbf{Your Task:}
        \begin{verbatim}
row 1: 'Name', 'Age', 'Occupation', 'City', 'Country'
row 2: 'Alice', '28', 'Engineer', 'Boston', 'USA'
row 3: 'Bob', '35', 'Teacher', 'London', 'UK'
row 4: 'Charlie', '42', 'Doctor', 'Sydney', 'Australia'
row 5: 'Diana', '31', 'Artist', 'Paris', 'France'
        \end{verbatim}
        Please create a properly formatted \{LATEX / MARKDOWN / TEXT\} table based on the information above.

        Your response:
    \end{tcolorbox}

    \begin{tcolorbox}[examplebox={outputbg}{Expect LATEX Table:}]
        \begin{verbatim}
<answer>
\begin{table*}[]
\begin{tabular}{|l|l|l|l|l|}
\hline
Name    & Age & Occupation & City        & Country   \\ \hline
Alice   & 28  & Engineer   & Boston      & USA       \\ \hline
Bob     & 35  & Teacher    & London      & UK        \\ \hline
Charlie & 42  & Doctor     & Sydney      & Australia \\ \hline
Diana   & 31  & Artist     & Paris       & France    \\ \hline
\end{tabular}
\end{table*}
</answer>
        \end{verbatim}
    \end{tcolorbox}

    \begin{tcolorbox}[examplebox={outputbg}{Expect MARKDOWN Table:}]
        \begin{verbatim}
<answer>
| Name    | Age | Occupation | City      | Country   |
|---------|-----|------------|-----------|-----------|
| Alice   | 28  | Engineer   | Boston    | USA       |
| Bob     | 35  | Teacher    | London    | UK        |
| Charlie | 42  | Doctor     | Sydney    | Australia |
| Diana   | 31  | Artist     | Paris     | France    |
</answer>
        \end{verbatim}
    \end{tcolorbox}
    
\end{tcolorbox}
\caption{Task examples of \textit{Aligned Table}.}
\label{fig:task_table1}
\end{figure*}

\begin{figure*}
\begin{tcolorbox}[taskbox, title=Aligned Table Case]

    \begin{tcolorbox}[examplebox={outputbg}{Expect ASCII Table:}]
        \begin{verbatim}
<answer>
+---------+-----+------------+-----------+-----------+
| Name    | Age | Occupation | City      | Country   |
+---------+-----+------------+-----------+-----------+
| Alice   | 28  | Engineer   | Boston    | USA       |
+---------+-----+------------+-----------+-----------+
| Bob     | 35  | Teacher    | London    | UK        |
+---------+-----+------------+-----------+-----------+
| Charlie | 42  | Doctor     | Sydney    | Australia |
+---------+-----+------------+-----------+-----------+
| Diana   | 31  | Artist     | Paris     | France    |
+---------+-----+------------+-----------+-----------+
</answer>
        \end{verbatim}
    \end{tcolorbox}

\end{tcolorbox}
\caption{Task examples of \textit{Aligned Table}.}
\label{fig:task_table2}
\end{figure*}

\subsection{Tree Understanding}
\label{app:task9}

\subsubsection*{Task Description}

This suite covers binary tree reasoning tasks based on ASCII-rendered trees, including structural question answering and path reconstruction between nodes.

\subsubsection*{Task Examples}
Figure~\ref{fig:task_tree} gives detailed task examples to both tree stucture and tree path tasks.

\begin{figure*}
\begin{tcolorbox}[taskbox, title=Tree Structure Case]
    \begin{tcolorbox}[examplebox={inputbg}{Input:}]
        You are given a binary tree structure. Please analyze it carefully and answer the question.

    \textbf{Tree Diagram Format:} Nodes are represented by numbers (1, 2, 3, etc.), "/" represents a connection to a left child, and "\\" represents a connection to a right child.

    Following are the Binary Tree Structure:

    \begin{verbatim}
                              115
                      /               \
              1                               89
          /       \                       /       \
      67              66              62              91
    /   \           /   \           /   \           /   \
  21      139     123     48      92      141     86      147
 / \     / \     / \     / \     / \     / \     / \     /
108 129 110 149 8   97  120 28  102 87  68  54  42  25  114
    \end{verbatim}
    \textbf{Question:} What is the right child of node 141?

    Put your final answer between <answer> and </answer> tags.
    \end{tcolorbox}
    
    \begin{tcolorbox}[examplebox={outputbg}{Expect Output:}]
<answer>54</answer>
    \end{tcolorbox}

\end{tcolorbox}

\begin{tcolorbox}[taskbox, title=Tree Path Case]
    \begin{tcolorbox}[examplebox={inputbg}{Input:}]
        You are given a binary tree structure. Please analyze it carefully and answer the question.

    \textbf{Tree Diagram Format:} Nodes are represented by numbers (1, 2, 3, etc.), "/" represents a connection to a left child, and "\\" represents a connection to a right child.

    Following are the Binary Tree Structure:

    \begin{verbatim}
              25
          /       \
      60              61
    /   \           /   \
  11      58      22      35
 / \     / \     / \
23  7   13  16  44  39
    \end{verbatim}
    \textbf{Question:} Find the path from node 35 to node 11 in the binary tree. Provide the path as a sequence of nodes separated by ' -> '.

    Put your final answer between <answer> and </answer> tags.
    \end{tcolorbox}
    
    \begin{tcolorbox}[examplebox={outputbg}{Expect Output:}]
<answer>35 -> 61 -> 25 -> 60 -> 11</answer>
    \end{tcolorbox}

\end{tcolorbox}
\caption{Task examples of \textit{Tree Understanding}.}
\label{fig:task_tree}
\end{figure*}

\subsection{ Map Navigation}
\label{app:task7}

\subsubsection*{Task Description}

This task measures spatial reasoning abilities on grid-based environments, including Sokoban and FrozenLake. Subtasks involve element identification (identify which element occupies given position), element location (determine the precise coordinates of specific object), surrounding elements (analyze and describe elements in adjacent cells), relative position (calculate relative spatial relationships between objects), and element counting (count specific element in the environment).

\subsubsection*{Task Examples}
Figure~\ref{fig:task_map} gives a detailed task example to Map Navigation task.
\begin{figure*}
    
\begin{tcolorbox}[taskbox, title=Map Navigation Case]
    \begin{tcolorbox}[examplebox={inputbg}{Input:}]
        You are an expert at spatial reasoning and map navigation. You will be given a Sokoban environment and asked questions about it.

    \textbf{Rules}: The environment is an N $\times$ N grid. Top-left corner is (0, 0), x-axis increases to the right (horizontal), y-axis increases downward (vertical). Each cell has a size of 1.

    Here is the Sokoban environment map:

    \begin{verbatim}
# # # # # # # # # # # #
# _ _ _ _ _ _ _ _ _ _ #
# _ _ # _ _ _ _ # P _ #
# _ # _ _ _ _ _ # # _ #
# _ _ # _ _ _ _ _ _ _ #
# _ _ _ _ _ _ _ _ _ _ #
# _ _ O # _ _ _ _ _ _ #
# _ _ _ _ # _ X _ # _ #
# _ _ _ _ _ _ _ _ _ _ #
# _ # _ _ _ _ # _ # _ #
# _ _ _ _ _ _ _ _ _ _ #
# # # # # # # # # # # #
    \end{verbatim}

    \textbf{Question:} What is the relative position from X to O? (Answer in format: (dx, dy))

    Please analyze the map and provide your answer. Remember to put your final answer between <answer> and </answer> tags.
    \end{tcolorbox}
    
    \begin{tcolorbox}[examplebox={outputbg}{Expect Output:}]
<answer>(-4, -1)</answer>
    \end{tcolorbox}

\end{tcolorbox}
\caption{A task example of \textit{Map Navigation}.}
\label{fig:task_map}
\end{figure*}

\subsection{ Gomoku State Reading}
\label{app:task6}

\subsubsection*{Task Description}

This benchmark evaluates board-reading and outcome classification on strictly generated Gomoku boards, each containing at most one winning line.

\subsubsection*{Task Examples}
Figure~\ref{fig:task_gomoku} gives a detailed task example to Gomoku tasks.
\begin{figure*}
\begin{tcolorbox}[taskbox, title=Gomoku Case]
    \begin{tcolorbox}[examplebox={inputbg}{Input:}]
        You are an expert at analyzing Gomoku (Five-in-a-Row) game boards. Your task is to determine the current game state by analyzing the board configuration.

    \textbf{Rules}: W represents a white stone, B represents a black stone, and E represents an empty position.

    Please analyze this Gomoku board and determine the game state, and wrap your final answer in <answer></answer> tags.
\begin{verbatim}
EBWWBEBEBWEWEWE
BBEEWWWEBEBWWWW
BEBWEBBBWBEBEEB
BBBWWBBEBWBBWEB
EWEWWWEBEBWEBBE
WWBWBEBWWBWEWWB
EWBEWBEWBWEEWWE
EEBBEBEEWWWWEWW
BEEWBEWBBEEBWEW
BWWBWEWEEWBWWWB
BWEBEBBEEWEEWEW
WWBWWBEWWWWBBBE
EBWBWWWWEWWBWWE
WWEEWWBWBEBBWEB
WBWBBBEEWBEWBWW
\end{verbatim}
    \end{tcolorbox}
    
    \begin{tcolorbox}[examplebox={outputbg}{Expect Output:}]
<answer>BLACK\_WINS</answer>
    \end{tcolorbox}

\end{tcolorbox}
\caption{A task example of \textit{Gomoku}.}
\label{fig:task_gomoku}
\end{figure*}


\subsection{RSA Difference}
\label{app:task8}

\subsubsection*{Task Description}

This benchmark evaluates the ability to localize and report all character-level differences between paired RSA fingerprint randomart patterns.

\subsubsection*{Task Examples}
Figure~\ref{fig:task_rsa} gives a detailed task example to RSA Difference tasks.

\begin{figure*}
\begin{tcolorbox}[taskbox, title=RSA Difference Case]
    \begin{tcolorbox}[examplebox={inputbg}{Input:}]
        You are given two RSA key fingerprint patterns. Your task is to find all the differences between them.

    \textbf{Rules}: Origin (0, 0) is at the top-left '+' character, x-axis increases to the right (horizontal), y-axis increases downward (vertical). Each character position has a length of 1.

    Following are the two RSA key fingerprint patterns for comparison:

    \begin{verbatim}
Pattern 1:                          Pattern 2:
+-[ RSA 2048 ]-+                    +-[ RSA 2048 ]-+
|=-^^oo.XOoX@*=|                    |=-^^oo.XOoX@*=|
|X#*=&*+@&%-=*#|                    |X#*=&*+@&%-=*#|
| +&^@Bo-O-^ -+|                    | +& @Bo-O-^ -+|
| X##O&=@=o@^-*|                    | X##O&=@=o@^-*|
|@+o%#@%%. @^ +|                    |@+o%#@%%. @^ +|
|^= BoO..o*.&#&|                    |^= BoO..o*.&#&|
|=%@= @-X#-&o+X|                    |=%@= @-X*-&o+X|
|*=@=*O=%^.#..-|                    |*=@=*O=%^.#..B|
+--------------+                    +--------------+
    \end{verbatim}
    Please identify all differences between Pattern 1 and Pattern 2. Put your final answer between <answer> and </answer> tags.
    \end{tcolorbox}
    
    \begin{tcolorbox}[examplebox={outputbg}{Expect Output:}]
<answer>
\begin{verbatim}
(4, 3): ^ ->  
(9, 7): # -> *
(14, 8): - -> B
\end{verbatim}
</answer>
    \end{tcolorbox}

\end{tcolorbox}
\caption{A task example of \textit{RSA Difference}.}
\label{fig:task_rsa}
\end{figure*}

\section{Benchmark Metrics and Task Generation}

This section outlines the detailed metrics for evaluating model outputs in \benchmarkname, as well as the generation process for each task. Unless specified otherwise, each sample is evaluated independently. Reported metrics represent macro‑averages across all samples, with further breakdowns by task‑specific attributes such as difficulty tier, format, or subtask type.

\subsection{Benchmark Metrics}
\label{app:metrics}

\paragraph{Answer Extraction.}
All tasks instruct the model to return the final response wrapped in \texttt{<answer>} and \texttt{</answer>} tags. We extract the first matched span between these tags as the \emph{extracted answer}. If either tag is missing, the extraction fails and the sample is counted as incorrect for exact-match based metrics, and 0 for both string similarity and F1 score. In each task, $p_i$ is the extracted model prediction and $g_i$ is the ground-truth answer.

\paragraph{Exact Match.}
For tasks scored by exact match, we compute:
\begin{equation}
\mathrm{EM} = \frac{1}{N}\sum_{i=1}^{N} \mathbb{I}\{p_i=g_i\}
\end{equation}

\paragraph{String Similarity.}
For tasks that also report average similarity, we use:
\begin{equation*}
\mathrm{Sim}(p_i,g_i) = 1 - \frac{d_{\mathrm{lev}}(p_i,g_i)}{\max(|p_i|,|g_i|)} 
\end{equation*}
\begin{equation}
    \mathrm{Avg. Sim} = \frac{1}{N}\sum_{i=1}^{N} \mathrm{Sim}(p_i,g_i)
\end{equation}
where $|·|$ stands for the sequence length, and $d_{\mathrm{lev}}$ is the Levenshtein distance.

\paragraph{F1 Score.}
For tasks where F1 score is applicable (Context-Aware Redaction and RSA Randomart Difference), we compute the Avg. F1 score:

\begin{equation*}
\mathrm{Precision}(p_i, g_i) = \frac{|p_i \cap g_i|}{|p_i|} 
\end{equation*}
\begin{equation}
\mathrm{Recall}(p_i, g_i) = \frac{|p_i \cap g_i|}{|g_i|}
\end{equation}
\begin{equation*}
\mathrm{F1}(p_i, g_i) = 2 \cdot \frac{\mathrm{Precision}(p_i, g_i) \cdot \mathrm{Recall}(p_i, g_i)}{\mathrm{Precision}(p_i, g_i) + \mathrm{Recall}(p_i, g_i)}
\end{equation*}
\begin{equation}
    \mathrm{Avg. F1} = \frac{1}{N}\sum_{i=1}^{N} \mathrm{F1}(p_i, g_i)
\end{equation}

Specifically, in the safety task, F1 score is computed at the number level, treating each sensitive value as a binary classification sample (redacted or not). The F1 score is then computed using precision and recall based on these counts.

In the RSA Randomart Difference task, F1 score is calculated based on the coordinate-level matching between predicted and ground truth differences. Precision and recall are computed for coordinate accuracy (matching x, y positions) and replacement accuracy (correctly predicted original and modified values at those coordinates). The coordinate F1 score and replacement F1 score are then computed separately. In the Table ~\ref{tab:benchmark_summary_final}, the coordinate F1 score is reported.

\paragraph{Content Score and Alignment Rate.}
The metrics, content score and alignment rate, are reported in Aligned Table task.

\subparagraph{Content Score.} The content score measures how accurately the model's generated table (e.g., \LaTeX, Markdown, or plain text) matches the ground truth answer, focusing on the precision of the table's cell contents. For each cell in the extracted table, it compares the predicted value against the corresponding ground truth value:

\begin{equation}
\text{Content Score} = \frac{1}{N}\sum_{i, j}  \mathbb{I}\{p_{i,j}=g_{i,j}\}
\end{equation}
where $N$ is the total cell number, $p_{i,j}$ is the cell content in extracted model prediction, and $g_i$ is the cell content in ground-truth answer.

\subparagraph{Alignment Rate.} The alignment rate measures how well the structure and formatting of the model's generated table match the expected table format (e.g., \LaTeX, Markdown, or plain text). It evaluates whether the table’s formatting (such as cell delimiters, line breaks, etc.) conforms to ground-truth answer.

The alignment rate is computed as the ratio of correctly aligned tables (where the structure and formatting match exactly) to the total number of evaluated tables:

\begin{equation}
\text{Alignment Rate} = \frac{\text{Number of Aligned Tables}}{\text{Total Number of Tables}}
\end{equation}

If the model produces exactly aligned tables in all cases, the alignment rate will be 1.0. A lower value indicates a lower level of alignment with the expected format.

\paragraph{Token Usage.}
When token logs are available, we report the traced completion token usage.

\begin{table*}[H]
\centering
\caption{Summary of Evaluation Metrics by Task.}
\label{tab:evaluation_metrics}
\resizebox{0.55\linewidth}{!}{%
\begin{tabular}{cl}
\toprule
\textbf{Benchmark Metric} & \textbf{Task} \\
\midrule
\textbf{Exact Match} & All tasks \emph{except for} Table Suite and RSA Difference \\
\textbf{Token Usage} & All tasks \\
\midrule
\textbf{Content Score} & Table Suite\\
\textbf{Alignment Rate} & Table Suite\\
\textbf{F1 Score} & Safety-style Masking and RSA Difference \\
\bottomrule
\end{tabular}
}
\end{table*}

\subsection{Task Generation}
\label{app:task_gen}

All tasks are generated through an automated test-generation pipeline, which can utilize a set of test parameters for controlled difficulty generation.

\paragraph{Data Source and Template.} Some tasks take the advantage of specific data source or template to generate contents. For the keystroke task, we use English words curated from \textbf{SCOWLv2} \footnote{\url{https://github.com/en-wl/wordlist}}. For the safety-style masking task, GPT-4.1 is used to generate content based on human-crafted, practical scenario templates. For cipher and decipher tasks, OCR-noise canonicalization, and table-related tasks, GPT-4.1 is employed to generate relevant content.

\paragraph{Pipeline and difficulty Control}
Each task-generation pipeline includes several parameters to control difficulty levels:

\begin{table*}[H]
\centering
\caption{Summary of Task Difficulty Control Parameters}
\label{tab:task_difficulty}
\resizebox{0.7\linewidth}{!}{%
\begin{tabular}{cl}
\toprule
\textbf{Benchmark Task} & \textbf{Difficulty Control Parameter} \\
\midrule
\textbf{Keystroke-level Text Editing} & Number of characters and types of manipulations \\
\textbf{Cipher \& Decipher} & Length of the original context \\
\textbf{Biological Sequence Manipulation} & Length of the biological sequence \\
\textbf{OCR-noise} & Length of the original context and the perturbation ratio \\
\textbf{Safety Mask} & Length of the original context and the number of sensitive numbers \\
\textbf{Table} & Number of cells in the table ($N \times M$) \\
\textbf{Tree} & Depth of the tree \\
\textbf{RSA Difference} & Size of the RSA pattern and the number of differences \\
\textbf{Map Navigation} & Size of the map \\
\textbf{Gomoku} & Size of the board and the density of the pieces \\
\bottomrule
\end{tabular}
}
\end{table*}

\section{Experiments setup}
\label{app:experiments}

\subsection{Benchmark Evaluation Setup}
\label{app:benchmark_setup}
\subsubsection{Model Configuration}
\label{app:model_setup}
\paragraph{Large-scale Models.} We evaluate large-scale models using their respective APIs with $max\_tokens$ set to the maximum allowed value. For models supporting temperature adjustment, we set $T=0.6$.

\paragraph{Local Models.} We perform inference using \textbf{vLLM} with the following settings for Qwen \& DS-distill-Qwen local models: $top\_p = 0.95$, temperature $T=0.6$, and $max\_tokens=32768$.

\subsubsection{Test Set Composition}
\label{app:test_setup}
The composition of test set in \benchmarkname{} evaluation result Table~\ref{tab:benchmark_summary_final} is listed in Table~\ref{tab:test_composition} below:

\begin{table*}[H]
\centering
\caption{\benchmarkname{} Test Set Composition}
\label{tab:test_composition}
\resizebox{0.8\linewidth}{!}{%
\begin{tabular}{llr}
\toprule
\textbf{Task} & \textbf{Parameters} & \textbf{Num} \\
\midrule
\textbf{Keystroke-level Text Editing (Encode)} & english word length $>10$ & 100 \\
\textbf{Keystroke-level Text Editing (Decode)} & token length $\in[16, 25]$, num\_backspaces $\in[6, 10]$ & 100 \\
\textbf{Cipher \& Decipher} & original context words $\in[150, 300]$ & 100 \\
\textbf{Biological Sequence Manipulation} & biological sequence length $=50$ & 100 \\
\textbf{OCR-noise} & original context words $\in[150, 300]$, perturbed ratio $=0.9$ & 100 \\
\textbf{Safety Mask} & sensitive numbers $\in[6, 10]$ & 100 \\
\textbf{Table} & $N\times M=8\times8$ & 100 \\
\textbf{Tree (Tree Structure)} & tree level $=6$ & 100 \\
\textbf{Tree (Tree Path)} & tree level $=6$ & 100 \\
\textbf{RSA Difference} & number of differences $=3, 5, 7, 9$ & each 100 \\
\textbf{Map Navigation (Sokoban)} & $N\times M=12\times12$ & 100 \\
\textbf{Map Navigation (FrozenLake)} & $N\times M=12\times12$ & 100 \\
\textbf{Gomoku (Linear Case)} & $N\times M=15\times15$, board density $=0.7$ & 100 \\
\textbf{Gomoku (Diagonal Case)} & $N\times M=15\times15$, board density $=0.7$ & 100 \\
\midrule
\textbf{Total} & & \textbf{1700} \\
\bottomrule
\end{tabular}
}
\end{table*}

\subsection{Number Linear Probe}
\label{app:linear_probe}
\paragraph{Goal.}
The aim of number linear probe is to study where a pretrained language model encodes \emph{character-count information} about an input string.

\paragraph{Label construction.}
Let $\mathcal{A}=\{a_1,\ldots,a_M\}$ be a dataset-specific alphabet of size $M=|\mathcal{A}|$,
and let $g:\Sigma^\ast\rightarrow\Sigma^\ast$ be a preprocessing function (e.g., lowercasing or
Unicode confusable normalization).
For an input string $w$, we define its bag-of-characters count vector
\begin{equation*}
y(w)\in\mathbb{N}^{M}
\quad
y_m(w)\;=\;\sum_{c\in g(w)} \mathbf{1}\!\left[c=a_m\right] \\
\end{equation*}
\begin{equation}
    m\in\{1,\ldots,M\}
\end{equation}
Let
\begin{equation}
K \;=\; \max_{w\in\mathcal{D}} \max_{m\in\{1,\ldots,M\}} y_m(w)
\end{equation}
denote the maximum count observed in the dataset $\mathcal{D}$.

\paragraph{Representations.}
Let $\mathrm{LM}$ be a frozen pretrained causal language model with $L$ layers.
Given a tokenized and padded version of $w$, let $t(w)$ denote the index of the \emph{\textbf{final} non-padding token}.
We extract, for each layer $\ell\in\{0,1,\ldots,L\}$, the \textbf{last-token} hidden state
\begin{equation}
h_\ell(w) \;=\; H_\ell(w)_{t(w)} \in \mathbb{R}^{d}
\end{equation}
where $H_\ell(w)\in\mathbb{R}^{T\times d}$ is the layer-$\ell$ hidden-state sequence
($\ell=0$ correspond to the embedding layer), $T$ is the padded sequence length,
and $d$ is the hidden size.

\paragraph{Linear probing.}
For each layer $\ell$, we train a \textbf{linear probe} that predicts the per-character counts from $h_\ell(w)$.
We model each character count as a $(K+1)$-way classification:
\begin{equation}
z_{\ell,m}(w) \;=\; W_{\ell,m} h_\ell(w) + b_{\ell,m}\in\mathbb{R}^{K+1}
\end{equation}
where $W_{\ell,m}\in\mathbb{R}^{(K+1)\times d}$ and $b_{\ell,m}\in\mathbb{R}^{K+1}$ are learned parameters.
The predicted count for character $a_m$ is
\begin{equation*}
\hat{y}_m(w) \;=\; \arg\max_{k\in\{0,\ldots,K\}} \; z_{\ell,m}(w)_k
\end{equation*}
\begin{equation}
    \hat{y}(w) = \bigl(\hat{y}_1(w),\ldots,\hat{y}_M(w)\bigr)
\end{equation}

\paragraph{Objective.}
Let $\alpha_{m,k}>0$ denote a class-imbalance weight for character $a_m$ and count class $k\in\{0,\ldots,K\}$.
We optimize a weighted cross-entropy objective over all characters and all samples:
\begin{equation*}
\mathcal{L}_\ell
\;=\;
\mathbb{E}_{w\sim\mathcal{D}}
\left[
\frac{1}{M}\sum_{m=1}^{M}
\alpha_{m,y_m(w)}
\cdot
Z_{\ell,m}
\right]
\end{equation*}
\begin{equation}
    Z_{\ell,m}=
-\log \bigl[\mathrm{softmax}\!\left(z_{\ell,m}(w)\right)\bigr]_{y_m(w)}
\end{equation}
We train $f_\ell=\{(W_{\ell,m},b_{\ell,m})\}_{m=1}^M$ independently for each layer $\ell$ while keeping $\mathrm{LM}$ frozen.

\paragraph{Metrics.}
We evaluate layer-wise decodability using: \textbf{Macro-averaged F1}, computed per character and then averaged across characters detailed as below:

For each sample $i\in\{1,\dots,N\}$, let the ground-truth count vector be
$y^{(i)}\in\{0,1,\dots,K\}^{M}$ and the probe prediction be
$\hat{y}^{(i)}\in\{0,1,\dots,K\}^{M}$. For each character $a_m\in\mathcal{A}$, define the confusion matrix
$C^{(m)}\in \mathbb{N}^{V\times V}$ whose entries count how often the true count is $t$ and the predicted count is $p$:
\begin{equation}
C^{(m)}_{t,p}
\;=\;
\sum_{i=1}^{N} \mathbf{1}\!\left[y^{(i)}_m=t \;\wedge\; \hat{y}^{(i)}_m=p\right]
\quad
t,p\in\{0,\dots,K\}
\end{equation}

For each class $c\in\{0,\dots,K\}$, define:
\begin{equation*}
\mathrm{TP}_{m,c} \;=\; C^{(m)}_{c,c}
\end{equation*}
\begin{equation*}
\mathrm{Pred}_{m,c} \;=\; \sum_{t=0}^{K} C^{(m)}_{t,c}
\end{equation*}
\begin{equation}
\mathrm{True}_{m,c} \;=\; \sum_{p=0}^{K} C^{(m)}_{c,p}
\end{equation}

Following the implementation, precision and recall are computed \emph{per class}:
\begin{equation}
\mathrm{Prec}_{m,c}
\;=\;
\frac{\mathrm{TP}_{m,c}}{\mathrm{Pred}_{m,c}}
\quad
\mathrm{Rec}_{m,c}
\;=\;
\frac{\mathrm{TP}_{m,c}}{\mathrm{True}_{m,c}}
\end{equation}

Then the per-character macro precision and macro recall are the uniform averages over classes:
\begin{equation}
P_m
\;=\;
\frac{1}{V}\sum_{c=0}^{K}\mathrm{Prec}_{m,c}
\quad
R_m
\;=\;
\frac{1}{V}\sum_{c=0}^{K}\mathrm{Rec}_{m,c}
\qquad
V=K+1
\end{equation}

The per-character F1 is computed from $P_m$ and $R_m$, and \textbf{macro-averaged F1} $F1_{\mathrm{macro}}$ is the uniform averages over characters:
\begin{equation}
F1_m
\;=\;
\begin{cases}
\dfrac{2P_mR_m}{P_m+R_m} & \text{if } P_m+R_m>0\\[8pt]
0 & \text{otherwise}
\end{cases}
\end{equation}
\begin{equation}
F1_{\mathrm{macro}}
\;=\;
\frac{1}{M}\sum_{m=1}^{M} F1_m
\end{equation}

\paragraph{Experiment configurations.}
We conduct number linear probing on Qwen2.5-7B-Instruct, keeping all base-model parameters frozen. We probe \emph{all} layers (layer $0$ to layer $28$, including the embedding output and every transformer block), and we always extract the hidden representation at the \textbf{final non-padding token} position.

Our \textbf{Normal dataset} contains $N=39{,}831$ English words (cleaned from \textbf{SCOWLv2} \footnote{\url{https://github.com/en-wl/wordlist}}, a database on English words). \textbf{Perturbed dataset} replaces each ASCII character in the normal dataset with a Unicode confusable counterpart with per-character probability $0.9$ (the same in the \textbf{OCR-noise canonicalization} task); \textbf{Random dataset} replaces each word with a uniformly sampled lowercase string of identical length (over \texttt{a--z}); and \textbf{Special dataset} replaces each word with a uniformly sampled string of identical length from the symbol alphabet
\texttt{\_PGO\#Xo+=.B*-@\%\&\^}.
For every dataset variant, we additionally run a \textbf{shuffle baseline} in which the extracted representations are randomly permuted across samples to break input--label alignment, yielding $8$ total runs.

We use the same train/test split for all runs with a $0.9/0.1$ ratio, seeded by 20250315.
For each layer, we train an independent \textbf{linear probe} (depth \(=1\)) with AdamW for $200$ epochs, batch size $8192$, and learning rate $10^{-3}$.

\begin{table*}[H]
\centering
\caption{Number linear probe experiment configurations.}
\label{tab:number_probe_config}
\resizebox{0.7\linewidth}{!}{%
\begin{tabular}{cl}
\toprule
\textbf{Component} & \textbf{Setting} \\
\midrule
\textbf{Base model} & \textsc{Qwen2.5-7B-Instruct} (frozen) \\
\textbf{Probed layers} & All layers $\ell \in \{0,\ldots,28\}$ (layer 0 for embedding) \\
\textbf{Representation} & Hidden state at the \textbf{final non-padding token} position \\
\midrule
\textbf{Train/test split} & $0.9/0.1$ \\
\textbf{Probe} & Linear probe (depth $=1$), trained independently per layer \\
\textbf{Optimizer} & AdamW \\
\textbf{Epochs} & $200$ \\
\textbf{Batch size} & $8192$ \\
\textbf{Learning rate} & $10^{-3}$ \\
\textbf{Random Seed} & $20250315$ \\
\bottomrule
\end{tabular}
}
\end{table*}

\subsection{Test-Time Budget Control (TTBC)}
\label{app:ttbc}
\paragraph{Goal.}
Our goal is to study \emph{test-time budget control} (TTBC) for reasoning models by \emph{explicitly controlling the length of the model's thinking trace} at inference time, and to quantify how task accuracy changes as a function of the enforced thinking-token budget.

\paragraph{Formulation.}
Each evaluation instance consists of an input prompt $x$ and (when available) a reference answer $y$.
We prompt the model to produce a response that decomposes into a \emph{thinking segment} $z$ and a \emph{final answer segment} $a$.
Let $\tau(\cdot)$ denote the tokenizer mapping text to a token sequence.
The \emph{thinking-token budget} is the length of the thinking segment in tokens:
\begin{equation}
T \;\triangleq\; |\tau(z)|
\end{equation}

Formally, given $x$, the model (with parameters $\theta$) produces a variable-length thinking trace
\begin{equation}
z \;=\; (z_1,\ldots,z_T)
\end{equation}
followed by an answer $a$.
A \emph{TTBC controller} $\mathcal{C}$ specifies an intervention rule that induces (i) a \emph{stopping time} $T_{\mathcal{C}}(x)$ for the thinking phase and (ii) a controlled distribution over thinking traces.
We denote the resulting controlled thinking distribution by $p_{\theta,\mathcal{C}}(z_{1:T}\mid x)$, where $T = T_{\mathcal{C}}(x)$.
After the controller terminates the thinking phase, the answer is generated conditional on the prompt and the realized thinking trace:
\begin{equation}
z_{1:T} \sim p_{\theta,\mathcal{C}}(z_{1:T}\mid x)
\qquad
a \sim p_{\theta}(a \mid x, z_{1:T})
\end{equation}

For each run, the realized thinking-token count $T \;=\; |\tau(z)|$, the answer-token count $A \;=\; |\tau(a)|$, 
and task performance computed from the extracted answer content are recorded.

\paragraph{Experiment configurations.}
We evaluate TTBC mechanisms by \textbf{Exact Thinking Tokens}, which enforces a strict thinking-token budget $t_{\mathrm{exact}}$.
If the model attempts to stop before reaching $t_{\mathrm{exact}}$, the controller injects the continuation cue (\textbf{\textit{"Wait"}}) to elicit more thinking; if the model exceeds the remaining budget, the thinking trace is truncated such that the final thinking length satisfies $T = t_{\mathrm{exact}}$. After the thinking segment is terminated by the TTBC rule, the system transitions to answer generation process.

To be specific, we run TTBC on the \textbf{Biological sequence manipulation} task (with sequence length set to 20), evaluating a fixed set of exact thinking budgets
$t_{\mathrm{exact}} \in \{256, 512, 1024, 2048, 4096, 8192, 16384\}$ tokens. We report \textbf{length-normalized similarity score} $r_{\mathrm{norm}}$ based on the Levenshtein distance $d_{\mathrm{lev}}(\hat{y}, y)$:
\begin{equation}
r_{\mathrm{norm}}(\hat{y}, y)
\;\triangleq\;
1 - \frac{d_{\mathrm{lev}}(\hat{y}, y)}{\max\{|\hat{y}|, |y|\}}
\end{equation}
where $|\cdot|$ denotes sequence length.

\begin{table*}[H]
\centering
\caption{TTBC experiment configurations on biological sequence manipulation.}
\label{tab:ttbc_config}
\resizebox{0.7\linewidth}{!}{%
\begin{tabular}{cl}
\toprule
\textbf{Component} & \textbf{Setting} \\
\midrule
\textbf{Model} & \textsc{DeepSeek-R1-Distill-Qwen-7B} \\
\textbf{Thinking segment} & \texttt{ <think> </think> } \\
\textbf{Answer segment} & \texttt{ <answer> <answer> } \\
\textbf{Continuation cue} & \textit{``Wait''} \\
\textbf{Thinking budget} & $ \{256,512,1024,2048,4096,8192,16384\}$ tokens \\
\midrule
\textbf{Temperature} & $0$ \\
\textbf{Samples} & $100$ \\
\textbf{Think length limit} & 32000 \\
\textbf{Answer length limit} & 2048 \\
\textbf{Random seed} & $0$ \\
\bottomrule
\end{tabular}
}
\end{table*}

\section{Supplementary Experiment Results}
\label{app:others}

\subsection{Details of Reasoning Model Budget}
\label{app:reasoning_budget}

\begin{figure*}[thbp]
    \centering
    
    \includegraphics[width=1\textwidth]{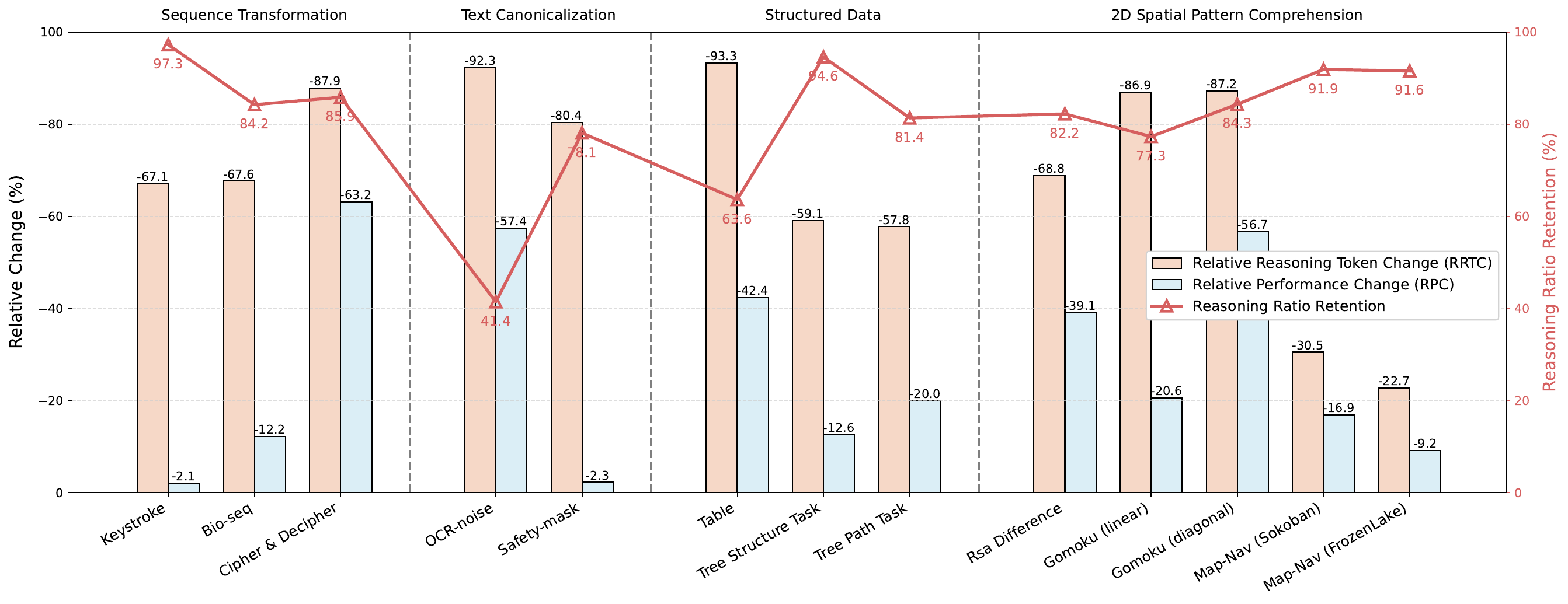}
    
    \caption{Relative changes in reasoning tokens, performance, and reasoning ratio retention between o4-mini (high) and o4-mini (low) across benchmark tasks. Bars show relative reductions in reasoning tokens (tan) and performance (blue), while the red line indicates reasoning ratio retention.}
    \label{fig:o4mini}
\end{figure*}

We list the detailed comparison of the performance of o4-mini (low) and o4-mini (high) in Figure~\ref{fig:o4mini} and Table~\ref{tab:o4mini_high_low_summary}. 

We define three key metrics to quantify the trade-offs: (1) \textbf{Relative Reasoning Token Change (RRTC)} measures the percentage reduction in reasoning tokens when switching from high to low budget, calculated as $\text{RRTC} = (\text{tokens}_{\text{high}} - \text{tokens}_{\text{low}})/{\text{tokens}_{\text{high}}} \times 100\%$; (2) \textbf{Relative Performance Change (RPC)} quantifies the performance impact, computed as $\text{RPC} = (\text{perf}_{\text{high}} - \text{perf}_{\text{low}})/{\text{perf}_{\text{high}}} \times 100\%$; (3) \textbf{Reasoning Ratio Retention} indicates the proportion of reasoning capacity retained in the low budget setting, defined as $\text{RR}_{\text{low}}/{\text{RR}_{\text{high}}} \times 100\%$, where RR denotes the reasoning ratio.

For most tasks, decreasing reasoning tokens leads to an obvious decline in o4-mini's performance. However, we observe that in certain tasks, such as Keystroke and Safety-mask, reducing reasoning tokens does not result in a substantial performance drop. This suggests the presence of overthinking issue, where the model allocates more computational resources than necessary for optimal performance.

\begin{table*}[htbp]
    \centering
    \caption{\textbf{o4-mini (high/low) Performance, Avg. Completion Tokens, and Reasoning Ratio}}
    \label{tab:o4mini_high_low_summary}
    \renewcommand{\arraystretch}{1.25}
    \resizebox{0.95\textwidth}{!}{%
    \begin{tabular}{l|cc|cc|cc}
        \toprule
        \multirow{2}{*}{\textbf{Benchmark Task}} 
        & \multicolumn{2}{c|}{\textbf{Score}} 
        & \multicolumn{2}{c|}{\textbf{Avg. Completion Tokens}} 
        & \multicolumn{2}{c}{\textbf{Reasoning Ratio}} \\
        \cmidrule(lr){2-3}\cmidrule(lr){4-5}\cmidrule(lr){6-7}
        & \textbf{o4-mini (high)} & \textbf{o4-mini (low)}
        & \textbf{o4-mini (high)} & \textbf{o4-mini (low)}
        & \textbf{o4-mini (high)} & \textbf{o4-mini (low)} \\
        \midrule

        OCR-noise                     & 47\% & 20\% & 3325  &  619 & 75.3\% & 31.2\% \\
        Table                 & 59\% & 34\% & 23276 & 2436 & 94.6\% & 60.2\% \\
        Bio-seq           & 98\% & 86\% & 3768  & 1447 & 90.1\% & 75.9\% \\
        Cipher \& Decipher               & 38\% & 14\% & 24118 & 3404 & 98.3\% & 84.4\% \\
        Safety-mask       & 88\% & 86\% & 3563  &  895 & 93.1\% & 72.7\% \\
        Gomoku (linear)               & 97\% & 77\% & 10816 & 1827 & 97.0\% & 75.0\% \\
        Gomoku (diagonal)             & 97\% & 42\% & 15259 & 2310 & 95.7\% & 80.7\% \\
        Map-Nav(Sokoban)      & 89\% & 74\% & 1423 & 1076  &  71.6\%    &  65.8\%    \\
        Map-Nav(FrozenLake)   & 87\% & 79\%  & 890  & 751 &  60.4\%  &  55.3\%    \\
        Rsa-diff (Avg. F1 score) & 0.9456  & 0.5760  &  7237   & 2744  &  99.1\% & 81.5\% \\
        Tree (Tree Structure)                  & 95\% & 83\% & 3914  & 1694 & 94.3\% & 89.2\%    \\
        Tree (Tree Path)                  & 90\% & 72\% & 2003  & 1040 & 97.6\% & 79.4\%    \\
        Keystroke                    & 97\% & 95\% & 1997  &  676 & 98.7\% & 96\%   \\

        \bottomrule
    \end{tabular}%
    }
\end{table*}

\subsection{Detailed sub-Task evaluation}
\label{app:sub_task}
\begin{figure*}[thbp]
    \centering
    \includegraphics[width=0.9\linewidth]{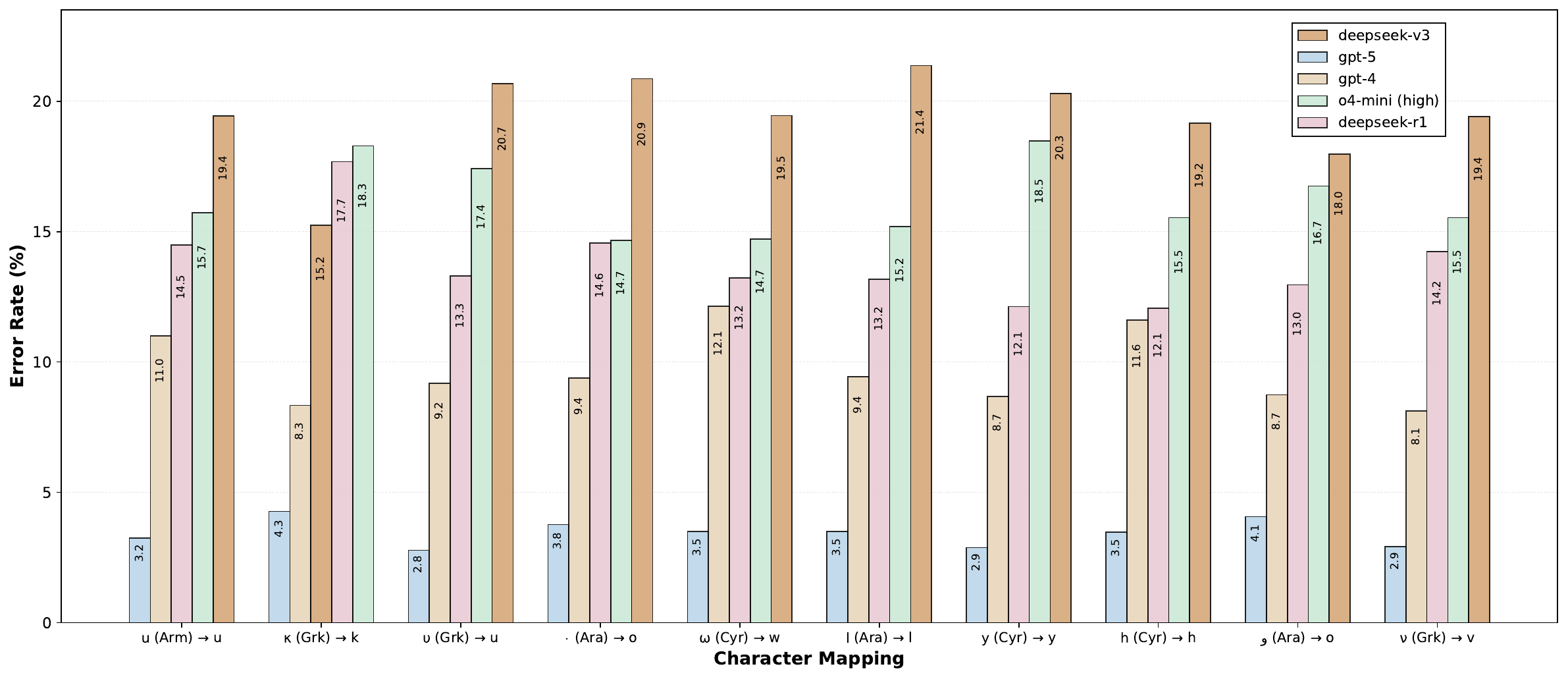}
    \caption{Worst mappings between normal and perturbed letters across
the large-scale models}
    \label{fig:mapping}
\end{figure*}

For the OCR-noise task, we aggregate the worst mappings between normal and perturbed letters across the large-scale models results in Figure~\ref{fig:mapping}.

In RSA difference task, we vary the number of differences in RSA patterns. In map navigation task, we evaluate performance across fine-grained sub-task types. Results for both tasks are provided in Table~\ref{tab:rsa_map_navigation}.

\begin{table*}[th]
    \centering
    \caption{Detailed performance on RSA Differences and Map Navigation Tasks. \small Abbreviations: DS-V3/R1 (DeepSeek-V3/R1), Ins (Instruct), DS-Qwen (DeepSeek-distill-Qwen).}
    \label{tab:rsa_map_navigation}
    \renewcommand{\arraystretch}{1.4}
    \resizebox{\textwidth}{!}{%
        \begin{tabular}{lll|cc|cc|cc|cccc}
            \toprule
            \multirow{2}{*}{\textbf{Group}} 
            & \multirow{2}{*}{\textbf{\makecell[l]{Benchmark \\ Task}}} 
            & \multirow{2}{*}{\textbf{Metric}} 
            & \multicolumn{2}{c|}{\textbf{DeepSeek}} 
            & \multicolumn{2}{c|}{\textbf{GPT}} 
            & \multicolumn{2}{c|}{\textbf{o}}
            & \multicolumn{4}{c}{\textbf{Qwen \& DS-distill-Qwen}} \\
            \cmidrule(lr){4-5} \cmidrule(lr){6-7} \cmidrule(lr){8-9} \cmidrule(l){10-13}
            & & 
            & \textbf{DS-V3} 
            & \textbf{DS-R1}
            & \textbf{GPT-4} 
            & \textbf{GPT-5} 
            & \textbf{\makecell{o4-mini\\(low)}} 
            & \textbf{\makecell{o4-mini\\(high)}}
            & \textbf{\makecell{Qwen-2.5\\-7B-Ins}} 
            & \textbf{\makecell{DS-Qwen-\\2.5-7B}} 
            & \textbf{\makecell{Qwen-2.5\\-32B-Ins}} 
            & \textbf{\makecell{DS-Qwen-\\2.5-32B}} \\
            \midrule

            
            \cellcolor{Spatial2D} & \multirow{7}{*}{\textbf{\makecell[l]{RSA-diff}}}
            & F1 (3 diff.) & 0.1075 & 0.9633 & 0.0687 & 0.9833 & 0.7799 & 0.9627 & 0.0169 & 0.0020 & 0.0198 & 0.0040 \\
            \cellcolor{Spatial2D} & & F1 (5 diff.) & 0.0549 & 0.9669 & 0.0378 & 0.9649 & 0.5699 & 0.9343 & 0.0047 & 0.0063 & 0.0073 & 0.0067 \\
            \cellcolor{Spatial2D} & & F1 (7 diff.) & 0.0697 & 0.9610 & 0.0314 & 0.9986 & 0.5013 & 0.9377 & 0.0065 & 0.0016 & 0.0156 & 0.0038 \\
            \cellcolor{Spatial2D} & & F1 (9 diff.) & 0.0641 & 0.9730 & 0.0385 & 0.9792 & 0.4528 & 0.9477 & 0.0082 & 0.0062 & 0.0079 & 0.0051 \\
            \hhline{~~-----------}
            \cellcolor{Spatial2D} & & Avg. F1 & 0.0741 & 0.9661 & 0.0441 & 0.9815 & 0.576 & 0.9456 & 0.0091 & 0.0040 & 0.0127 & 0.0049 \\
            \cellcolor{Spatial2D} & & \cellcolor{lightgray}Token Num & \cellcolor{lightgray}1331 & \cellcolor{lightgray}12665 & \cellcolor{lightgray}79 & \cellcolor{lightgray}6624 & \cellcolor{lightgray}2744 & \cellcolor{lightgray}7237 & \cellcolor{lightgray}494 & \cellcolor{lightgray}1496 & \cellcolor{lightgray}99 & \cellcolor{lightgray}1251 \\
            \hhline{~------------}
            
            \cellcolor{Spatial2D} & \multirow{6}{*}{\textbf{\makecell[l]{Map-Nav \\ (Sokoban)}}}
            & Element ID Acc. & 100.0\% & 84.0\% & 92.0\% & 100.0\% & 80.0\% & 76.0\% & 96.0\% & 0.0\% & 80.0\% & 40.0\% \\
            \cellcolor{Spatial2D} & & Element Loc Acc. & 88.0\% & 100.0\% & 40.0\% & 100.0\% & 72.0\% & 96.0\% & 12.0\% & 36.0\% & 40.0\% & 96.0\% \\
            \cellcolor{Spatial2D} & & Surrounding Elem Acc. & 56.0\% & 96.0\% & 8.0\% & 92.0\% & 84.0\% & 96.0\% & 0.0\% & 0.0\% & 12.0\% & 72.0\% \\
            \cellcolor{Spatial2D} & & Relative Pos Acc. & 76.0\% & 100.0\% & 52.0\% & 100.0\% & 60.0\% & 88.0\% & 8.0\% & 8.0\% & 16.0\% & 68.0\% \\
            \hhline{~~-----------}
            \cellcolor{Spatial2D} & & Avg. Acc. & 79.0\% & 99.0\% & 48.0\% & 98.0\% & 74.0\% & 89.0\% & 29.0\% & 11.0\% & 37.0\% & 69.0\% \\
            \cellcolor{Spatial2D} & & \cellcolor{lightgray}Token Num & \cellcolor{lightgray}331 & \cellcolor{lightgray}1478 & \cellcolor{lightgray}44 & \cellcolor{lightgray}891 & \cellcolor{lightgray}1076 & \cellcolor{lightgray}1423 & \cellcolor{lightgray}41 & \cellcolor{lightgray}2839 & \cellcolor{lightgray}119 & \cellcolor{lightgray}1114 \\
            \hhline{~------------}
            
            \cellcolor{Spatial2D} & \multirow{7}{*}{\textbf{\makecell[l]{Map-Nav \\ (FrozenLake)}}}
            & Element ID Acc. & 100.0\% & 100.0\% & 80.0\% & 100.0\% & 100.0\% & 100.0\% & 90.0\% & 25.0\% & 85.0\% & 75.0\% \\
            \cellcolor{Spatial2D} & & Element Loc Acc. & 90.0\% & 100.0\% & 55.0\% & 95.0\% & 85.0\% & 90.0\% & 5.0\% & 35.0\% & 35.0\% & 90.0\% \\
            \cellcolor{Spatial2D} & & Surrounding Elem Acc. & 85.0\% & 95.0\% & 60.0\% & 100.0\% & 85.0\% & 95.0\% & 5.0\% & 15.0\% & 55.0\% & 75.0\% \\
            \cellcolor{Spatial2D} & & Relative Pos Acc. & 90.0\% & 100.0\% & 25.0\% & 100.0\% & 40.0\% & 50.0\% & 0.0\% & 10.0\% & 20.0\% & 75.0\% \\
            \cellcolor{Spatial2D} & & Count Elem Acc. & 100.0\% & 100.0\% & 75.0\% & 100.0\% & 85.0\% & 100.0\% & 65.0\% & 5.0\% & 40.0\% & 85.0\% \\
            \hhline{~~-----------}
            \cellcolor{Spatial2D} & & Avg. Acc. & 93.0\% & 99.0\% & 59.0\% & 99.0\% & 79.0\% & 87.0\% & 33.0\% & 18.0\% & 47.0\% & 80.0\% \\
            \cellcolor{Spatial2D}\multirow{-20}{*}{\rotatebox[origin=c]{90}{\color{black}\textbf{\makecell{2D Spatial Pattern\\Comprehension}}}} & & \cellcolor{lightgray}Token Num & \cellcolor{lightgray}289 & \cellcolor{lightgray}1010 & \cellcolor{lightgray}23 & \cellcolor{lightgray}602 & \cellcolor{lightgray}751 & \cellcolor{lightgray}890 & \cellcolor{lightgray}21 & \cellcolor{lightgray}1918 & \cellcolor{lightgray}96 & \cellcolor{lightgray}1052 \\
            
            \bottomrule
        \end{tabular}
    }
\end{table*}

\section{Error Analysis}

\subsection{Benchmark Tasks Error Analysis}
\label{app:benchmark_error}

\subsubsection*{Tokenization-induced Errors}
Tokenization-induced errors occur when the model's tokenization process interferes with its ability to accurately process character-level or position-specific information. These errors are particularly prevalent in \benchmarkname tasks, and Figure~\ref{fig:token1}, Figure~\ref{fig:token2}, Figure~\ref{fig:token3}, Figure~\ref{fig:token4}, and Figure~\ref{fig:token5} are some examples in \benchmarkname{}.

\begin{figure*}[!htbp]
    \begin{blackbox}{Response from o4-mini (high)}
\textbf{Tree Structure Task}
\tiny 
\begin{verbatim}
                                                              86
                                              /                               \
                              161                                                             112
                      /               \                                               /               \
              100                             182                             142                             223
          /       \                       /       \                       /       \                       /       \
      229             39              97              155             85              22              222             123
    /   \           /   \           /   \           /   \           /   \           /   \           /   \           /   \
  240     177     246     91      276     169     267     96      195     206     251     69      262     82      265     232
 / \     / \     / \     / \     / \     /         \       \     / \     / \       \     / \     / \     / \     / \     /
45  167 152 219 176 2   124 242 125 163 27          110     46  178 65  70  133     158 227 55  146 23  204 9   140 218 54
\end{verbatim}
\normalsize
\begin{verbatim}
Question: What is the right child of node 262?
\end{verbatim}
\textbf{Model Response:}
\small

Here’s how the bottom‐level children line up under each node at level 4 (240, 177, 246, 91, 276, 169, 267, 96, 195, 206, 251, 69, 262, 82, 265, 232):

240$\rightarrow$45,167  

177$\rightarrow$152,219  

246$\rightarrow$176,2  

91$\rightarrow$124,242  

276$\rightarrow$125,163  

169$\rightarrow$27  

267$\rightarrow$110  

96$\rightarrow$46  

195$\rightarrow$178,65  

206$\rightarrow$70,133  

\textcolor{red}{251$\rightarrow$158,227}  

\textcolor{red}{69$\rightarrow$55,146}  

\textcolor{red}{262$\rightarrow$23,204}  

\textcolor{red}{82$\rightarrow$9,140}  

\textcolor{red}{265$\rightarrow$218,54}  

\textcolor{red}{232$\rightarrow$None}  

Thus, the right child of 262 is 204.

<answer>\textcolor{red}{204}</answer>

\normalsize

\textbf{Ground Truth:}

<answer>23</answer>

    \end{blackbox}
\caption{Tokenization-induced error example in Tree Understanding task.}
\label{fig:token1}
\end{figure*}

\begin{figure*}[!htbp]
\begin{blackbox}{Response from deepseek-v3}
\textbf{Tree Structure Task}
\tiny 
\begin{verbatim}
                                                            198
                                            /                               \
                            175                                                             33
                    /               \                                               /               \
            82                              20                              210                             147
        /       \                       /       \                       /       \                       /       \
    200             228             69              137             212             91              142             78
  /   \           /   \           /   \           /   \           /   \           /   \           /   \           /   \
171     60      50      136     251     266     223     274     221     120     43      169     48      230     119     238
 \     / \     / \     / \     / \     / \     / \     / \     /       / \     / \     / \     /       / \
  84  201 54  121 176 246 118 277 153 111 76  247 117 216 215 31      213 61  275 101 72  47  49      5   244
\end{verbatim}
\normalsize
\begin{verbatim}
Question: What is the parent node of node 275?
\end{verbatim}
\textbf{Model Response:}
\small

Let's trace the path to node 275 in the given tree.

Starting from root 198:

- 198 $\rightarrow$ left child 175 $\rightarrow$ right child 20 $\rightarrow$ left child 69 $\rightarrow$ right child 266 $\rightarrow$ left child 251 $\rightarrow$ right child 101 $\rightarrow$ left child 275?  

Wait, check carefully:  

Actually, 69 has left child 251 and right child 266.  

266 has left child 277 and right child 153? No, the diagram shows:  
\begin{verbatim}
69  
/  \  
251 266  
/ \  / \  
... / \ ...  
\end{verbatim}
Wait, let's read the tree carefully:

From 69:  
Left child 251 $\rightarrow$ children: 277, 153  

Right child 266 $\rightarrow$ children: 111, 76  

So 275 is not under 69.

Check 147 subtree:  

147 $\rightarrow$ left child 142 $\rightarrow$ right child 230 $\rightarrow$ \textcolor{red}{left child 72 $\rightarrow$ right child 47? No, 230 has children: 119, 238.}  

So not there.

Check 33 subtree:  

33 $\rightarrow$ left child 210 $\rightarrow$ left child 212 $\rightarrow$ left child 221 $\rightarrow$ right child 31? No.  

33 $\rightarrow$ right child 147 $\rightarrow$ left child 142 $\rightarrow$ left child 48 $\rightarrow$ right child 49? No.

Check 175 subtree:  

175 $\rightarrow$ left child 82 $\rightarrow$ right child 228 $\rightarrow$ left child 50 $\rightarrow$ left child 121 \textcolor{red}{$\rightarrow$ right child 176?}

Check 33 $\rightarrow$ left child 210 $\rightarrow$ right child 91 $\rightarrow$ \textcolor{red}{left child 43 $\rightarrow$ right child 169 $\rightarrow$ left child 275?}  
Let's verify:  
\begin{verbatim}
91  
/  \  
43  169  
/ \  / \  
213 61 275 101  
\end{verbatim}

\textcolor{red}{Yes! 169 has left child 275 and right child 101.}  

So parent of 275 is 169.

<answer>\textcolor{red}{169}</answer>

\normalsize

\textbf{Ground Truth:}

<answer>43</answer>

\end{blackbox}
\caption{Tokenization-induced error example in Tree Understanding task.}
\label{fig:token2}
\end{figure*}

\begin{figure*}[!htbp]
\begin{blackbox}{Response from gpt-4}
\textbf{Safety-mask Task}
\small

You are a data privacy assistant. Your task is to redact (mask) sensitive information in the given text according to specific rules.

**Redaction Rules:**

1. **18-digit ID Card Number**: Keep the first 6 digits and last 2 digits, mask the middle 10 digits with asterisks (*)

   - Example: 123456789012345678  123456**********78

2. **Phone Number** (format: +[area code] [11-digit number]): Keep '+' sign and area code, keep first 3 digits and last 4 digits of phone number, mask the middle 4 digits with asterisks (*)

   - Example: +12 12345678901  +12 123****8901

3. **Credit Card Number**: Keep the first 6 digits and last 4 digits, mask all middle digits with asterisks (*)

   - Example: 1234 5678 9012 3456  123456******3456

**Text to redact:**

In a recent customer record audit, we identified discrepancies in the contact details provided. The primary contact number listed for Mr. 703324375459458270 does not match our records. We attempted to reach out using the alternative +82 87901400470 provided but were unable to establish contact. It is crucial to update this information promptly to ensure seamless communication going forward. Additionally, we noticed that the credit card on file expired last month. We kindly request that Mr. 620140087033621286 updates the payment method to avoid any disruptions to their service.
During the investigation into the discrepancies in the records, we discovered that the 963871589151698749 was not updated in our system following the recent renewal. This oversight may have contributed to the inconsistencies in the provided contact information. To rectify this, we urge Mr. 727095999340105436 to provide us with the updated details at their earliest convenience. Moreover, we noticed that the credit card ending in 7194 2835 7809 3796 is nearing its expiration date. We advise Mr. 842012233125116498 to update the payment information to prevent any delays in processing upcoming payments. Thank you for your prompt attention to these matters.

**Your redacted text (place between <answer></answer> tags):**

\normalsize
\textbf{Model Response:}
\small

<answer>In a recent customer record audit, we identified discrepancies in the contact details provided. The primary contact number listed for Mr. \textcolor{red}{703324******59270} does not match our records. We attempted to reach out using the alternative +82 879****0470 provided but were unable to establish contact. It is crucial to update this information promptly to ensure seamless communication going forward. Additionally, we noticed that the credit card on file expired last month. We kindly request that Mr. \textcolor{red}{620140******1286} updates the payment method to avoid any disruptions to their service.
During the investigation into the discrepancies in the records, we discovered that the \textcolor{red}{963871******9749} was not updated in our system following the recent renewal. This oversight may have contributed to the inconsistencies in the provided contact information. To rectify this, we urge Mr. \textcolor{red}{727095******5436} to provide us with the updated details at their earliest convenience. Moreover, we noticed that the credit card ending in 719428******3796 is nearing its expiration date. We advise Mr. \textcolor{red}{842012******6498} to update the payment information to prevent any delays in processing upcoming payments. Thank you for your prompt attention to these matters.</answer>

\normalsize

\textbf{Evaluation Result:}
\begin{verbatim}
Incorrectly Redacted/Missing (5):
     id_card: 703324375459458270  703324**********70 [incorrect_or_missing]
     id_card: 620140087033621286  620140**********86 [incorrect_or_missing]
     id_card: 963871589151698749  963871**********49 [incorrect_or_missing]
     id_card: 727095999340105436  727095**********36 [incorrect_or_missing]
     id_card: 842012233125116498  842012**********98 [incorrect_or_missing]
\end{verbatim}

\end{blackbox}
\caption{Tokenization-induced error example in Safety-mask task.}
\label{fig:token3}
\end{figure*}

\begin{figure*}[!htbp]
\begin{blackbox}{Response from deepseek-r1}
\textbf{Safety-mask Task}
\small

You are a data privacy assistant. Your task is to redact (mask) sensitive information in the given text according to specific rules.

**Redaction Rules:**

1. **18-digit ID Card Number**: Keep the first 6 digits and last 2 digits, mask the middle 10 digits with asterisks (*)

   - Example: 123456789012345678  123456**********78

2. **Phone Number** (format: +[area code] [11-digit number]): Keep '+' sign and area code, keep first 3 digits and last 4 digits of phone number, mask the middle 4 digits with asterisks (*)

   - Example: +12 12345678901  +12 123****8901

3. **Credit Card Number**: Keep the first 6 digits and last 4 digits, mask all middle digits with asterisks (*)

   - Example: 1234 5678 9012 3456  123456******3456

**Text to redact:**

In the latest customer records review, we identified a discrepancy in +81 65885951847 associated with 791203559727501899 for account number 626578496720551032. It seems like there was a mix-up during the data entry process, leading to incorrect information being linked to the account. This issue needs urgent attention to ensure accurate customer details, especially considering the upcoming audit next month. The team is working on rectifying the error and updating the phone number linked to 342091894702652726 to reflect the correct information. 
As part of the security protocol, please ensure that any changes made to customer records, such as updating +86 41971093480 or 588853786584622252, are documented properly and verified by a supervisor to avoid any data integrity issues. For any further inquiries regarding this matter or assistance needed in updating the information, please reach out to the respective department heads. We value data accuracy and privacy highly, and it is essential to maintain strict confidentiality when handling sensitive information like 829629097998353799 and +86 65088382784. Your cooperation in ensuring proper data management practices is greatly appreciated.

**Your redacted text (place between <answer></answer> tags):**

\normalsize
\textbf{Model Response:}
\small

<answer>In the latest customer records review, we identified a discrepancy in +81 658****1847 associated with 791203**********99 for account number 626578**********32. It seems like there was a mix-up during the data entry process, leading to incorrect information being linked to the account. This issue needs urgent attention to ensure accurate customer details, especially considering the upcoming audit next month. The team is working on rectifying the error and updating the phone number linked to 342091**********26 to reflect the correct information. 
As part of the security protocol, please ensure that any changes made to customer records, such as updating +86 419****3480 or \textcolor{red}{588853**********22}, are documented properly and verified by a supervisor to avoid any data integrity issues. For any further inquiries regarding this matter or assistance needed in updating the information, please reach out to the respective department heads. We value data accuracy and privacy highly, and it is essential to maintain strict confidentiality when handling sensitive information like 829629**********99 and +86 650****2784. Your cooperation in ensuring proper data management practices is greatly appreciated.</answer>

\normalsize

\textbf{Evaluation Result:}
\begin{verbatim}
Incorrectly Redacted/Missing (1):
     id_card: 588853786584622252  588853**********52 [incorrect_or_missing]
\end{verbatim}

\end{blackbox}
\caption{Tokenization-induced error example in Safety-mask task.}
\label{fig:token4}
\end{figure*}

\begin{figure*}[!htbp]
\begin{blackbox}{Response from deepseek-r1}
\textbf{Keystroke Task}
\small

You are a typewriter effect simulator. Your task is to show the progressive typing sequence of a given word.

TASK DESCRIPTION:
Given an input word, you need to generate the complete sequence showing how the word appears character by character as if someone is typing it.

RULES:
1. Start with the first character
2. Add one character at a time until the complete word is formed
3. Use "$\leftarrow$" (arrow) to separate each step
4. Do not include any extra characters, spaces, or explanations
5. Only output the progressive sequence

IMPORTANT:
- Your response should ONLY contain the progressive sequence
- No explanations, no additional text
- Follow the exact format shown in examples

INPUT: maintainability

\normalsize
\textbf{Model Response:}

<answer>m$\leftarrow$ma$\leftarrow$mai$\leftarrow$main$\leftarrow$maint\textcolor{red}{$\leftarrow$mainta$\leftarrow$maintain}$\leftarrow$maintaina$\leftarrow$maintainab$\leftarrow$maintainabi
$\leftarrow$maintainabil$\leftarrow$maintainabili$\leftarrow$maintainabilit$\leftarrow$maintainability</answer>

\textbf{Ground Truth:}

</answer>m$\leftarrow$ma$\leftarrow$mai$\leftarrow$main$\leftarrow$maint$\leftarrow$mainta$\leftarrow$maintai$\leftarrow$maintain$\leftarrow$maintaina$\leftarrow$maintainab
$\leftarrow$maintainabi$\leftarrow$maintainabil$\leftarrow$maintainabili$\leftarrow$maintainabilit$\leftarrow$maintainability</answer>

\end{blackbox}
\caption{Tokenization-induced error example in Keystroke task.}
\label{fig:token5}
\end{figure*}

\subsubsection*{Overthinking}

Overthinking manifests when models generate excessively verbose reasoning chains that provide minimal analytical benefit while consuming substantial computational resources. This behavior is particularly pronounced in reasoning models, which often produce convoluted explanations for straightforward problems. 

In practice, overthinking appears as lengthy step-by-step analyses that either circle redundant points or introduce unnecessary complexity. Figure~\ref{fig:overthink1} and Figure~\ref{fig:overthink2} are some examples in \benchmarkname{}.

\begin{figure*}[!htbp]
\begin{blackbox}{Response from deepseek-r1}
\textbf{RSA Difference Task}
\small

\begin{verbatim}

You are given two RSA key fingerprint patterns. Your task is to find all the differences between them.

Following are the two RSA key fingerprint patterns for comparison:
+-------------[ RSA 2048 ]-------------+  +-------------[ RSA 2048 ]-------------+
|.OO=..XooX#^B&%-^*X+%o=o.+X-^=&*+--^B#|  |.OO=..XooX#^B&%-^*X+%o=o.+X-^=&*+--^B#|
|+B=&B*^.^+o**+^*#XX+*X-#*+OoO.B%#.*XOX|  |+B=&B*^.^+o**+^*#XX+*X-#*+OoO.B%#.*XOX|
|O*oo.+&. .#B=@O-@@o#=+..X+-&O-o=.%@ &#|  |O*oo.+&. .#B=@O-@@o#=+..X+-&O-o=.%@ &#|
|B#&+X&+@^.@.O&OX^@oB@o%-B-- o&oX* #o&#|  |B#&+X&+@^.@.O&OX^@oB@o%-B-- o&oX* #o&#|
|+&*XX+B=*%=o.X-oB@%%^*=-=OB%@^X@# @*.%|  |+&*XX+B=*%=o.X-oB@O%^*=-=OB%@^X@# @*.%|
|+B-.B @+-#=@%X+=%oB*#oBX%++%XB* X+=-X&|  |+B-.B @+-#=@%X+=%oB*#oBX%++%XB* X+=-X&|
|B%&o%+^ %% +O@^o@+O%*#=#@ &^*^.*^=^XBX|  |B%&o%+^ %% +O@^o@+O%*#=#@ &^*^.*^=^XBX|
|.o. &B+% .X.*BB&#.X*O&@X.BX%&*%^# *@#-|  |.o. &B+% .X.*BB&#.X*O&@X.BX%&*%^# *@#-|
|%X^B -B.-.=&&OX==% #oo^-*#*#Oo#*@^@#oO|  |%X^B -B.-.=&&OX==% #oo^-*#*#Oo#*@^@#oO|
|+*+-.*B +.B%X %& o&o ^*^+-o*o-B^*OX@.&|  |+*+-.*B +.B%X %& o&o ^*^+-o*o-B^*OX@.&|
|##X ^@#*OO.OB X@o+.X +.^oB*=BX.&*XO@@&|  |##X ^@#*OO.OB X@o+.X +.^oB*=BX.&*XO@@&|
|*X&@^O#-@.^@o=*o.%+@BBOo+^+^O^B*%@-^  |  |*X&@^O#-@.^@o=*o.%+@BBOo+^+^O^B*%@-^  |
|=^@oB -Xo @O. .*&-o-B@##o&=^o=^@ XB=.@|  |=^@oB -Xo @O. .*&-o-B@##o&=^o=^@ XB=.@|
|-=^^-OOO&^O+^@X^X@#+@+XO+oo@oXO=-%@B@@|  |-=^^-OOO&^O+^@X^X@#+@+XO+oo@o^O=-%@B@@|
|&oBB B.= +^@ ^ -@-X.*-*#=B-+OBo^oOo#o+|  |&oBB B.= +^@ ^ -@-X.*-*#=B-+OBo^oOo#o+|
|#&O%+  ^*oOo#B*O-.*XB.^*@&=.=@OB*-X&B+|  |#&O%+  ^*oOo#B*O-.*XB.^*@&=.=@OB*-X&B+|
|%+^=%O%&@%@+. B+o @X*++%^oX@*@#o.=#O =|  |%+^=%O%&@%@+. B+  @X*++%^oX@*@#o.=#O =|
|X-B@--B=&@B.O..+XX%@+@o+ .X +Oo.X=-O #|  |X-B@--B=&@B.O.O+XX%@+@o+ .X %Oo.X=-O #|
+--------------------------------------+  +--------------------------------------+
\end{verbatim}
\normalsize
\textbf{Model Thinking:}
\small
We are given two patterns, Pattern 1 and Pattern 2. We need to compare them character by character, noting any differences. The coordinate system has origin at the top-left '+' character of each pattern. The patterns are given as lines of text. 

Wait, that's only 40 characters? Actually, I counted up to index 39. Let's count the string length properly. 

Let's write it out:

\textcolor{blue}{[Thinking Process omit]}

So indeed 40 characters. But the top border has 43. That means the box is not exactly aligned? \textcolor{red}{Wait}, the top border line is: "+-------------[ RSA 2048 ]-------------+"

\textcolor{blue}{[Thinking Process omit]}

So indeed, there are 40 characters: indices 0 to 39. That means the left bar is at x=0 and the right bar is at x=39. But the top border has a '+' at x=0 and x=42. 

\textcolor{red}{Wait}, if we count the total length, it is 40 characters? Let's count manually:
Write the string:

\textcolor{blue}{[Thinking Process omit]}

Actually, Pattern1 x=14 is '.' and Pattern2 x=14 is also '.'? Wait, in Pattern2, after "O." there is "O", so the characters are: x=12 '.', x=13 'O', x=14 '.', x=15 'O'. So at x=14, both are '.'. The difference is at x=15: Pattern1 has '.', Pattern2 has 'O'.
\textcolor{red}{So difference at (15,18): . -> O.}

\textcolor{blue}{[Thinking Process omit]}

So it is 41 characters? Actually, from index 0 to 40: that's 41 characters. But I counted up to index 40: that gives 41 characters. 
\textcolor{red}{Wait}, but my list above has indices 1 to 40? Let's do 0-based indexing:

\textcolor{blue}{[Thinking Process omit]}

That is 40 characters? Actually, indices 0 to 39 give 40 characters. But index 39 is '+'? In my list, index 39 is '+'? \textcolor{red}{Wait}, index 38 would be the 39th character. So total indices 0 to 38: 39 characters? I need to be precise.

\textcolor{blue}{[Thinking Process omit]}

\textcolor{red}{Wait}, in Pattern2, after "O." there is "O", so the characters are: x=12 '.', x=13 'O', x=14 '.', x=15 'O'. \textcolor{red}{So at x=15, both are '.'}. 
Also, later in the line:

\textcolor{blue}{[Thinking Process omit]}

But \textcolor{red}{wait}, are there any other lines? We checked y=1 to y=18. We should also check y=0 and y=19, but they are borders and identical.

\textcolor{blue}{[Thinking Process omit]}

<answer>
(19,5): \% -> O
(30,14): X -> \^
(17,17): o ->
(29,18): + -> \%
</answer>

But \textcolor{red}{wait}: in the example format, they had spaces around the arrow? 

\textcolor{blue}{[Thinking Process omit]}

So I'll write exactly as shown.

Thus, final answer.

\normalsize

\textbf{Model Response:}

<answer>
(19, 5): \% -> O
(30, 14): X -> \^
(17, 17): o ->  
(29, 18): + -> \%
</answer>

\textbf{Ground Truth:}

(29, 18): + -> \%
\textcolor{red}{(15, 18): . -> O}
(30, 14): X -> \^
(19, 5): \% -> O
(17, 17): o ->  

\end{blackbox}
\caption{Overthinking example in RSA Difference task.}
\label{fig:overthink1}
\end{figure*}

\begin{figure*}[!htbp]
\begin{blackbox}{Response from deepseek-r1}
\textbf{Gomoku Task}
\small

\begin{verbatim}

You are an expert at analyzing Gomoku (Five-in-a-Row) game boards. 
Your task is to determine the current game state by analyzing the board configuration.

Rules:
- W represents a white stone
- B represents a black stone  
- E represents an empty position
- Victory condition: 5 consecutive stones of the same color in a row (horizontal, vertical, 
or diagonal)
- Only one player can win, or no one wins

You should analyze the board carefully and provide your reasoning. 
Then, wrap your final answer in <answer></answer> tags with exactly one of these three options:
- WHITE_WINS if white has 5 consecutive stones
- BLACK_WINS if black has 5 consecutive stones
- NO_WINNER if neither player has achieved victory

Please analyze this Gomoku board and determine the game state:

Board size: 15x15
                                    EWWWEEWWWBBWWBW
                                    BEBBWEWEBWEBBBW
                                    WEBWWBBEBWWWWBE
                                    EWEBBEEBEBBBEWE
                                    WWBEBWEBWWEWEEB
                                    BWBEWBWWBWWWEWW
                                    WBBBBWEBWBWWBBE
                                    EWWBEWBEWWWBWBW
                                    BEBWBBBBWWEBEEW
                                    BEBWBEWEEEEWWBB
                                    BWEBWEBWBWWEWBB
                                    EWWBEWBBEWEEBEW
                                    BEBWWWEBEWWWWEB
                                    BWWBBEEBEWWEBEE
                                    EBWBBBBEWBBWWWW
    
What is the current game state?
\end{verbatim}
\normalsize
\textbf{Model Thinking:}
\small

First, I need to analyze this Gomoku board....

\textcolor{blue}{[Thinking Process omit]}

\textcolor{red}{Row 13: B E B W W W W E B E W W W W E B? Wait, the row is given as: BEBWWWEBEWWWWEB}

\textcolor{red}{Let me write it carefully: Row 13: B E B W W W W E B E W W W W E B? But the string is "BEBWWWEBEWWWWEB" which has 15 characters.}

\textcolor{red}{Indices: 0:B, 1:E, 2:B, 3:W, 4:W, 5:W, 6:W, 7:E, 8:B, 9:E, 10:W, 11:W, 12:W, 13:W, 14:E? No, from the string: BEBWWWEBEWWWWEB — let's list: B,E,B,W,W,W,W,E,B,E,W,W,W,W,E,B? That would be 16 characters? I think I miscounted.}

\textcolor{red}{The board has 15 rows, each with 15 characters. So Row 13 must be 15 characters. The user wrote: "BEBWWWEBEWWWWEB" – let's count: B E B W W W W E B E W W W W E B – that's 16 characters? B,E,B,W,W,W,W,E,B,E,W,W,W,W,E,B — yes, 16. There must be a mistake.}

I need to check the input...

\textcolor{blue}{[Thinking Process omit]}

\textcolor{red}{Row 13: BEBWWWEBEWWWWEB -> let's count: B E B W W W W E B E W W W W E B -> that's 16 characters? B, E, B, W, W, W, W, E, B, E, W, W, W, W, E, B -> 16. But it should be 15. Perhaps there's a typo.} Let's see the user's input: "BEBWWWEBEWWWWEB" – if I take it as 15 characters, maybe it's "BEBWWWEBEWWWWE" or something? But from the context, I need to assume it's 15x15, so perhaps I misread.

\textcolor{blue}{[Thinking Process omit]}

\textcolor{red}{Line 13: "BEBWWWEBEWWWWEB" - let's count: B E B W W W W E B E W W W W E B -> if I count the letters: B=1, E=2, B=3, W=4, W=5, W=6, W=7, E=8, B=9, E=10, W=11, W=12, W=13, W=14, E=15, B=16?} That seems like 16. But in the string, there are no spaces, so "BEBWWWEBEWWWWEB" has 15 characters? Let's write it out: B E B W W W W E B E W W W W E B — that's 16 characters. Perhaps it's "BEBWWWEBEWWWWE" with 15? But the user wrote "BEBWWWEBEWWWWEB" which ends with B, so maybe it's 15 and I miscounted.

\textcolor{red}{Let's use a counter: B E B W W W W E B E W W W W E B}

\textcolor{blue}{[Thinking Process omit]}
\end{blackbox}
\end{figure*}

\begin{figure*}[!htbp]
\begin{blackbox}{}

Now, I'll count the characters in row13: "BEBWWWEBEWWWWEB" - let's write each character:

1:B, 2:E, 3:B, 4:W, 5:W, 6:W, 7:W, 8:E, 9:B, 10:E, 11:W, 12:W, 13:W, 14:W, 15:E, 16:B? 

\textcolor{blue}{[Thinking Process omit]}

Now, let's take row13: "BEBWWWEBEWWWWEB" - if I paste this into a text editor, it shows 15 characters? Let's count manually: B-1, E-2, B-3, W-4, W-5, W-6, W-7, E-8, B-9, E-10, W-11, W-12, W-13, W-14, E-15, B-16? No, after W-14, it is E and then B, but in the string, "EWWWWEB" might be considered as E, W, W, W, W, E, B which is 7 characters. So from start: B E B W W W W E B E W W W W E B -> that's 16. But perhaps "BEBWWWEB" is 8 characters? B,E,B,W,W,W,W,E,B? That's 9? I'm getting confused.

\textcolor{blue}{[Thinking Process omit]}

Row13: "BEBWWWEBEWWWWEB" - now, let's say this has 15 characters. So I need to interpret it as 15 characters. Let's list: B, E, B, W, W, W, W, E, B, E, W, W, W, W, E? 

\textcolor{blue}{[Thinking Process omit]}

I think I should proceed with row13 as "BEBWWWEBEWWWWE" for analysis. Otherwise, I can't continue.

\textcolor{red}{So for row13, I'll use: B E B W W W W E B E W W W W E}

\textcolor{blue}{[Thinking Process omit]}

So after thorough checking, I cannot find any five consecutive stones of the same color in any direction.

Therefore, the game state should be \textcolor{red}{NO\_WINNER.}

\normalsize

\textbf{Model Response:}

After analyzing the board configuration for horizontal, vertical, and diagonal sequences of five consecutive stones, no such sequence was found for either white (W) or black (B). Therefore, neither player has achieved victory.

<answer>\textcolor{red}{NO\_WINNER}</answer>

\textbf{Ground Truth:} BLACK\_WINS

\end{blackbox}
\caption{Overthinking example in Gomoku task.}
\label{fig:overthink2}
\end{figure*}

\subsection{TTBC Error Analysis}
\label{app:ttbc_error}

\subsubsection*{Increasing Phase}

As the number of thinking tokens $t_{exact}$ increases from 256 to 512, the model begins to engage in a more detailed thinking process before providing the final answer for biological sequence tasks. Figure~\ref{fig:pahse1} is an example from the DeepSeek-R1-Distill-Qwen-7B in the increasing phase.

\begin{figure*}[!htbp]
    \begin{blackbox}{Response from DS-distill-Qwen-2.5-7B with $t_{exact}$ 256}
\begin{verbatim}
<|im_start|>system
You are a helpful reasoning assistant.
<|im_end|>
<|im_start|>user
\end{verbatim}
\textcolor{blue}{[Instructions omitted]}
\begin{verbatim}
**Input DNA sequence:** GTAGTCGTGTGGTATTTTCG
Please provide the complementary DNA sequence for the given input.<|im_end|>
<|im_start|>assistant
<think>
Okay, so I need to find the complementary DNA sequence for the given input, which is 
GTAGTCGTGTGGTATTTTCG. Hmm, I remember that DNA is made up of four nucleotides...
...
Starting with the first base, G. Since G pairs with C, the first complement will be C.
Next is T. T pairs with A, so the second complement is A.
Then comes A. A pairs with T, so the third
\end{verbatim}
\textcolor{red}{[Thinking process is truncated by TTBC rule.]}

</think>

<answer>

CATCCGAGTTGTGCAACTGC

</answer>

\textbf{Expected answer: CAT}CAGCACACCATAAAAGC
    \end{blackbox}

\caption{Increasing phase example in biological sequence task.}
\label{fig:pahse1}
\end{figure*}

\subsubsection*{Plateau Phase}

When the thinking tokens, $t_{exact}$, range from 1024 to 2048, the model starts to perform double-checking after receiving the "\textbf{\textit{Wait}}" continuation cue. Figure~\ref{fig:pahse2} is an example from the DeepSeek-R1-Distill-Qwen-7B in the plateau phase.

\begin{figure*}[!htbp]
    \begin{blackbox}{Response from DS-distill-Qwen-2.5-7B with $t_{exact}$ 1024}
\begin{verbatim}
<|im_start|>system
You are a helpful reasoning assistant.
<|im_end|>
<|im_start|>user
\end{verbatim}
\textcolor{blue}{[Instructions omitted]}
\begin{verbatim}
**Input protein sequence (three-letter codes):** GLU-TYR-TRP-PRO-GLN-VAL-....

Please provide the one-letter protein sequence for the given input.<|im_end|>
<|im_start|>assistant
<think>
Okay, so I need to convert this protein sequence from three-letter codes to one-
letter codes. Let me see, the input is GLU-TYR-TRP-PRO-GLN-VAL-... 
\end{verbatim}
\textcolor{blue}{[Thinking process omitted]}
\begin{verbatim}
So, joining them without any separators, it should be EYWPQVSDFS.....

Wait, let me double-check each conversion to make sure I didn't make a mistake. 
GLU to E, correct. TYR to Y, correct. TRP to W, correct. PRO to P, correct...
\end{verbatim}
\textcolor{red}{[Thinking process is truncated by TTBC rule.]}

</think>
<answer>

EYWPQVSDFSTAILSVFMMC

</answer>

\textbf{Expected answer:} EYWPQVSDFSTAILSVPFMM
    \end{blackbox}
\caption{Plateau phase example in biological sequence task.}
\label{fig:pahse2}
\end{figure*}

\subsubsection*{Decreasing Phase}

Once the thinking tokens, $t_{exact}$, exceed 2048, reaching values such as 4096 or 8192, the model begins to exhibit mistakes due to overthinking. Figure~\ref{fig:pahse3} is an example from the DeepSeek-R1-Distill-Qwen-7B in the decreasing phase.

\begin{figure*}[!htbp]
    \begin{blackbox}{Response from DS-distill-Qwen-2.5-7B with $t_{exact}$ 8192}
\begin{verbatim}
<|im_start|>system
You are a helpful reasoning assistant.
<|im_end|>
<|im_start|>user
\end{verbatim}
\textcolor{blue}{[Instructions omitted]}GCGAUCAGUG
\begin{verbatim}
**Input RNA sequence:** AUAACCAGAUGCGAUCAGUG

Please provide the complementary RNA sequence for the given input.<|im_end|>
<|im_start|>assistant
<think>
Okay, so I need to find the complementary RNA sequence for the given RNA sequence. 
Hmm, I remember that in RNA, the bases pair up in a specific way. Let me think...
\end{verbatim}
\textcolor{blue}{[Thinking process omitted]}
\begin{verbatim}
Wait, let me make sure I'm doing this right. Each base in the original RNA is 
replaced by its complement. So, for each position, I replace A with U, U with A, 
C with G, and G with C.
Let me go through each position step by step:
1. The first base is A. Its complement is U.
2. The second base is U. Its complement is A.
......
\end{verbatim}
\textcolor{blue}{[Thinking process omitted]}
\begin{verbatim}
Wait, let me double-check to make sure I didn't mix up any bases....
So the complementary sequence is U A U U G G U C U A C G C U A G U C A C
......
\end{verbatim}
\textcolor{blue}{[Thinking process omitted]}
\begin{verbatim}
Wait, but I'm a bit confused because sometimes I think the complementary sequence 
is written in reverse complement form, but in this case, since we're just finding 
the complementary sequence without reversing.

Wait, but I just thought of something else. Let me make sure I didn't mix up any 
bases. Let me list them again:
Original: U G A U A A C C A G A U ...
Complement: A C U A U U G G U C U A ...
......
\end{verbatim}
\textcolor{red}{[Thinking process is truncated by TTBC rule.]}

</think>

<answer>

\textcolor{red}{AC}UAUUGGUCUAC\textcolor{red}{C}CUAGUCAC\textcolor{red}{UU}

</answer>

\textbf{Expected answer:} UAUUGGUCUACGCUAGUCAC
    \end{blackbox}
\caption{Decreasing phase example in biological sequence task.}
\label{fig:pahse3}
\end{figure*}

\end{appendices}












    

\end{document}